\documentclass[sigconf]{acmart}
\pdfoutput=1

\usepackage{algorithm,algpseudocode,amsmath,booktabs,color,multirow,paralist,subfigure,xspace}

\acmConference[arXiv]{arXiv}{arXiv}{arXiv}
\acmYear{2017}
\newcommand{\superscript}[1]{\ensuremath{^{\textrm{#1}}}}
\def\myfirst{\superscript{1}}
\def\mysecond{\superscript{2}}

\newcommand*{\Scale}[2][4]{\scalebox{#1}{$#2$}}

\theoremstyle{definition}
\newtheorem*{definition*}{Definition}
\newtheorem*{problem*}{Problem}
\newtheorem*{example*}{Example}

\usepackage{etoolbox}
\patchcmd{\maketitle}{\@copyrightspace}{}{}{}

\begin{document}
\title{MetaPAD: Meta Pattern Discovery from Massive Text Corpora}

\author{Meng Jiang\myfirst, Jingbo Shang\myfirst, Taylor Cassidy\mysecond, Xiang Ren\myfirst}
\author{Lance M. Kaplan\mysecond, Timothy P. Hanratty\mysecond, Jiawei Han\myfirst}
\affiliation{
	\institution{{\myfirst}Department of Computer Science, University of Illinois Urbana-Champaign, IL, USA}
	\institution{{\mysecond}Computational \& Information Sciences Directorate, Army Research Laboratory, Adelphi, MD, USA}
	{\myfirst}{\{mjiang89, shang7, xren7, hanj\}@illinois.edu}
	{\mysecond}{\{taylor.cassidy.civ, lance.m.kaplan.civ, timothy.p.hanratty.civ\}@mail.mil}
}
\renewcommand{\authors}{Meng Jiang, Jingbo Shang, Taylor Cassidy, Xiang Ren,
	Lance M. Kaplan, Timothy P. Hanratty, and Jiawei Han}
\renewcommand{\shortauthors}{M. Jiang et al.}

\begin{abstract}
Mining textual patterns in news, tweets, papers, and many other kinds of text corpora has been an active theme in text mining and NLP research. 
Previous studies adopt a dependency parsing-based pattern discovery approach. 
However, the parsing results lose rich context around entities in the patterns, and the process is costly for a corpus of large scale. 
In this study, we propose a novel \emph{typed textual pattern structure}, called \emph{meta pattern}, which is extended to a frequent, informative, and precise subsequence pattern in certain context. 
We propose an efficient framework, called \textsf{MetaPAD}, which discovers meta patterns from massive corpora with three techniques:
(1) it develops a context-aware segmentation method to carefully determine the boundaries of patterns with a learnt pattern quality assessment function, which avoids costly dependency parsing and generates high-quality patterns; 
(2) it identifies and groups synonymous meta patterns from multiple facets---their types, contexts, and extractions; and
(3) it examines type distributions of entities in the instances extracted by each group of patterns, and looks for appropriate type levels to make discovered patterns precise.
Experiments demonstrate that our proposed framework discovers high-quality typed textual patterns efficiently from different genres of massive corpora and facilitates information extraction.

\end{abstract}

%\begin{CCSXML}
%<ccs2012>
%  <concept_id>10010520.10010553.10010562</concept_id>
%  <concept_desc>Computer systems organization~Embedded systems</concept_desc>
%  <concept_significance>500</concept_significance>
%  <concept_id>10010520.10010575.10010755</concept_id>
%  <concept_desc>Computer systems organization~Redundancy</concept_desc>
%  <concept_significance>300</concept_significance>
%  <concept_id>10010520.10010553.10010554</concept_id>
%  <concept_desc>Computer systems organization~Robotics</concept_desc>
%  <concept_significance>100</concept_significance>
%  <concept_id>10003033.10003083.10003095</concept_id>
%  <concept_desc>Networks~Network reliability</concept_desc>
%  <concept_significance>100</concept_significance>
%</ccs2012>
%\end{CCSXML}

%\ccsdesc[500]{Computer systems organization~Embedded systems}

% We no longer use \terms command

%\keywords{ACM proceedings, \LaTeX, text tagging}

\maketitle

\newcommand{\scxdocument}{$\mathcal{D}$\xspace}
\newcommand{\scxword}{$w$\xspace}
\newcommand{\scxphrase}{$p$\xspace}
\newcommand{\scxmark}{$m$\xspace}
\newcommand{\scxwordset}{$\mathcal{W}$\xspace}
\newcommand{\scxphraseset}{$\mathcal{P}$\xspace}
\newcommand{\scxmarkset}{$\mathcal{M}$\xspace}
\newcommand{\scxentity}{$e$\xspace}
\newcommand{\scxname}{$a$\xspace}
\newcommand{\scxvalue}{${v}_{i}$\xspace}
\newcommand{\scxvaluearray}{$\mathbf{v}$\xspace}
\newcommand{\scxclass}{\textsc{C}\xspace}
\newcommand{\scxnameset}{$\mathcal{A}$\xspace}
\newcommand{\scxvaluetype}{${T}_{i}$\xspace}
\newcommand{\scxvaluetypearray}{$\mathbf{T}$\xspace}
\newcommand{\scxentityset}{$\mathcal{E}$\xspace}
\newcommand{\scxallentityset}{$\mathcal{E}^{+}$\xspace}
\newcommand{\scxclassset}{$\mathcal{C}$\xspace}
\newcommand{\scxallclassset}{$\mathcal{C}^{+}$\xspace}
\newcommand{\scxhierarchy}{$\mathcal{H}$\xspace}
\newcommand{\scxmetapattern}{$MP$\xspace}

\newcommand{\scdocument}{$\mathcal{D}$}
\newcommand{\scword}{$w$}
\newcommand{\scphrase}{$p$}
\newcommand{\scmark}{$m$}
\newcommand{\scwordset}{$\mathcal{W}$}
\newcommand{\scphraseset}{$\mathcal{P}$}
\newcommand{\scmarkset}{$\mathcal{M}$}
\newcommand{\scentity}{$e$}
\newcommand{\scname}{$a$}
\newcommand{\scvalue}{${v}_{i}$}
\newcommand{\scvaluearray}{$\mathbf{v}$}
\newcommand{\scclass}{\textsc{C}}
\newcommand{\scnameset}{$\mathcal{A}$}
\newcommand{\scvaluetype}{${T}_{i}$}
\newcommand{\scvaluetypearray}{$\mathbf{T}$}
\newcommand{\scentityset}{$\mathcal{E}$}
\newcommand{\scallentityset}{$\mathcal{E}^{+}$}
\newcommand{\scclassset}{$\mathcal{C}$}
\newcommand{\scallclassset}{$\mathcal{C}^{+}$}
\newcommand{\schierarchy}{$\mathcal{H}$}
\newcommand{\scmetapattern}{$MP$}

\newcommand{\cslocation}{$\$$\textsc{Location}\xspace}
\newcommand{\csaqua}{$\$$\textsc{Aqua}\xspace}
\newcommand{\csocean}{$\$$\textsc{Ocean}\xspace}
\newcommand{\csriver}{$\$$\textsc{River}\xspace}
\newcommand{\cssea}{$\$$\textsc{Sea}\xspace}
\newcommand{\csgovloc}{$\$$\textsc{GovLoc}\xspace}
\newcommand{\cscity}{$\$$\textsc{City}\xspace}
\newcommand{\csstate}{$\$$\textsc{State}\xspace}
\newcommand{\cscnprovince}{$\$$\textsc{CNProvince}\xspace}
\newcommand{\cscountry}{$\$$\textsc{Country}\xspace}
\newcommand{\cscounty}{$\$$\textsc{County}\xspace}
\newcommand{\csethnicity}{$\$$\textsc{Ethnicity}\xspace}
\newcommand{\csusstate}{$\$$\textsc{USState}\xspace}
\newcommand{\csland}{$\$$\textsc{Land}\xspace}
\newcommand{\csisland}{$\$$\textsc{Island}\xspace}
\newcommand{\cspeninsula}{$\$$\textsc{Peninsula}\xspace}
\newcommand{\cspath}{$\$$\textsc{Path}\xspace}
\newcommand{\csavenue}{$\$$\textsc{Avenue}\xspace}
\newcommand{\csboulevard}{$\$$\textsc{Boulevard}\xspace}
\newcommand{\csdrive}{$\$$\textsc{Drive}\xspace}
\newcommand{\csroad}{$\$$\textsc{Road}\xspace}
\newcommand{\csstreet}{$\$$\textsc{Street}\xspace}
\newcommand{\cspoi}{$\$$\textsc{POI}\xspace}
\newcommand{\csairport}{$\$$\textsc{Airport}\xspace}
\newcommand{\csbridge}{$\$$\textsc{Bridge}\xspace}
\newcommand{\cshospital}{$\$$\textsc{Hospital}\xspace}
\newcommand{\cssquare}{$\$$\textsc{Square}\xspace}
\newcommand{\csstadium}{$\$$\textsc{Stadium}\xspace}
\newcommand{\csstation}{$\$$\textsc{Station}\xspace}
\newcommand{\csorganization}{$\$$\textsc{Organization}\xspace}
\newcommand{\cscompany}{$\$$\textsc{Company}\xspace}
\newcommand{\csairlines}{$\$$\textsc{Airlines}\xspace}
\newcommand{\csevent}{$\$$\textsc{Event}\xspace}
\newcommand{\csattack}{$\$$\textsc{Attack}\xspace}
\newcommand{\csaward}{$\$$\textsc{Award}\xspace}
\newcommand{\csfestival}{$\$$\textsc{Festival}\xspace}
\newcommand{\cssummit}{$\$$\textsc{Summit}\xspace}
\newcommand{\csgame}{$\$$\textsc{Game}\xspace}
\newcommand{\cspolicy}{$\$$\textsc{Policy}\xspace}
\newcommand{\csprotest}{$\$$\textsc{Protest}\xspace}
\newcommand{\cswar}{$\$$\textsc{War}\xspace}
\newcommand{\csgovernment}{$\$$\textsc{Government}\xspace}
\newcommand{\csinstitute}{$\$$\textsc{Institute}\xspace}
\newcommand{\csinternational}{$\$$\textsc{International}\xspace}
\newcommand{\csmilitantgroup}{$\$$\textsc{MilitantGroup}\xspace}
\newcommand{\csmovie}{$\$$\textsc{Movie}\xspace}
\newcommand{\csnewsagency}{$\$$\textsc{NewsAgency}\xspace}
\newcommand{\csparty}{$\$$\textsc{Party}\xspace}
\newcommand{\csproduct}{$\$$\textsc{Product}\xspace}
\newcommand{\csreligion}{$\$$\textsc{Religion}\xspace}
\newcommand{\cssportsleague}{$\$$\textsc{SportsLeague}\xspace}
\newcommand{\cssportsteam}{$\$$\textsc{SportsTeam}\xspace}
\newcommand{\csteamname}{$\$$\textsc{TeamName}\xspace}
\newcommand{\csperson}{$\$$\textsc{Person}\xspace}
\newcommand{\csartist}{$\$$\textsc{Artist}\xspace}
\newcommand{\csactor}{$\$$\textsc{Actor}\xspace}
\newcommand{\csauthor}{$\$$\textsc{Author}\xspace}
\newcommand{\csdirector}{$\$$\textsc{Director}\xspace}
\newcommand{\csmodel}{$\$$\textsc{Model}\xspace}
\newcommand{\cssinger}{$\$$\textsc{Singer}\xspace}
\newcommand{\cstvhost}{$\$$\textsc{TVHost}\xspace}
\newcommand{\cstvpersonality}{$\$$\textsc{TVPersonality}\xspace}
\newcommand{\csathlete}{$\$$\textsc{Athlete}\xspace}
\newcommand{\csbaseballplayer}{$\$$\textsc{BaseballPlayer}\xspace}
\newcommand{\csbasketballplayer}{$\$$\textsc{BasketballPlayer}\xspace}
\newcommand{\csboxer}{$\$$\textsc{Boxer}\xspace}
\newcommand{\csclimber}{$\$$\textsc{Climber}\xspace}
\newcommand{\cscyclist}{$\$$\textsc{Cyclist}\xspace}
\newcommand{\csfootballplayer}{$\$$\textsc{FootballPlayer}\xspace}
\newcommand{\cscornerback}{$\$$\textsc{CornerBack}\xspace}
\newcommand{\cslinebacker}{$\$$\textsc{LineBacker}\xspace}
\newcommand{\csquarterback}{$\$$\textsc{QuarterBack}\xspace}
\newcommand{\csrunningback}{$\$$\textsc{RunningBack}\xspace}
\newcommand{\cstightend}{$\$$\textsc{TightEnd}\xspace}
\newcommand{\cswidereceiver}{$\$$\textsc{WideReceiver}\xspace}
\newcommand{\csgolfer}{$\$$\textsc{Golfer}\xspace}
\newcommand{\cssoccerplayer}{$\$$\textsc{SoccerPlayer}\xspace}
\newcommand{\cstennisplayer}{$\$$\textsc{TennisPlayer}\xspace}
\newcommand{\csattacker}{$\$$\textsc{Attacker}\xspace}
\newcommand{\csmurderer}{$\$$\textsc{Murderer}\xspace}
\newcommand{\csterrorist}{$\$$\textsc{Terrorist}\xspace}
\newcommand{\csattorney}{$\$$\textsc{Attorney}\xspace}
\newcommand{\csbusinessperson}{$\$$\textsc{Businessperson}\xspace}
\newcommand{\cscharacter}{$\$$\textsc{Character}\xspace}
\newcommand{\cscoach}{$\$$\textsc{Coach}\xspace}
\newcommand{\csbasketballcoach}{$\$$\textsc{BasketballCoach}\xspace}
\newcommand{\csfootballcoach}{$\$$\textsc{FootballCoach}\xspace}
\newcommand{\csjournalist}{$\$$\textsc{Journalist}\xspace}
\newcommand{\csleader}{$\$$\textsc{Leader}\xspace}
\newcommand{\cspoliceofficer}{$\$$\textsc{PoliceOfficer}\xspace}
\newcommand{\cspolitician}{$\$$\textsc{Politician}\xspace}
\newcommand{\csattorneygeneral}{$\$$\textsc{AttorneyGeneral}\xspace}
\newcommand{\csforeignminister}{$\$$\textsc{ForeignMinister}\xspace}
\newcommand{\csgeneral}{$\$$\textsc{General}\xspace}
\newcommand{\csgovernor}{$\$$\textsc{Governor}\xspace}
\newcommand{\csjudge}{$\$$\textsc{Judge}\xspace}
\newcommand{\csmayor}{$\$$\textsc{Mayor}\xspace}
\newcommand{\cspresident}{$\$$\textsc{President}\xspace}
\newcommand{\csprimeminister}{$\$$\textsc{PrimeMinister}\xspace}
\newcommand{\csrepresentative}{$\$$\textsc{Representative}\xspace}
\newcommand{\cssecretaryofdefense}{$\$$\textsc{SecretaryOfDefense}\xspace}
\newcommand{\cssecretaryofstate}{$\$$\textsc{SecretaryOfState}\xspace}
\newcommand{\cssenator}{$\$$\textsc{Senator}\xspace}
\newcommand{\csspokesperson}{$\$$\textsc{Spokesperson}\xspace}
\newcommand{\csvicepresident}{$\$$\textsc{VicePresident}\xspace}
\newcommand{\csprotester}{$\$$\textsc{Protester}\xspace}
\newcommand{\csscientist}{$\$$\textsc{Scientist}\xspace}
\newcommand{\csvictim}{$\$$\textsc{Victim}\xspace}
\newcommand{\csminister}{$\$$\textsc{Minister}\xspace}
\newcommand{\csday}{$\$$\textsc{Day}\xspace}
\newcommand{\csdigit}{$\$$\textsc{Digit}\xspace}
\newcommand{\csdigitrank}{$\$$\textsc{DigitRank}\xspace}
\newcommand{\csdigitunit}{$\$$\textsc{DigitUnit}\xspace}
\newcommand{\csemail}{$\$$\textsc{Email}\xspace}
\newcommand{\cslengthunit}{$\$$\textsc{LengthUnit}\xspace}
\newcommand{\csmonth}{$\$$\textsc{Month}\xspace}
\newcommand{\csphone}{$\$$\textsc{Phone}\xspace}
\newcommand{\cstime}{$\$$\textsc{Time}\xspace}
\newcommand{\cstimeunit}{$\$$\textsc{TimeUnit}\xspace}
\newcommand{\csurl}{$\$$\textsc{URL}\xspace}
\newcommand{\csweekday}{$\$$\textsc{Weekday}\xspace}
\newcommand{\csyear}{$\$$\textsc{Year}\xspace}
\newcommand{\cstreatment}{$\$$\textsc{Treatment}\xspace}
\newcommand{\csdisease}{$\$$\textsc{Disease}\xspace}
\newcommand{\csbacteria}{$\$$\textsc{Bacteria}\xspace}
\newcommand{\csantibiotics}{$\$$\textsc{Antibiotics}\xspace}

\newcommand{\csuniversity}{$\$$\textsc{University}\xspace}
\newcommand{\csprofessor}{$\$$\textsc{Professor}\xspace}
\newcommand{\csresearcher}{$\$$\textsc{Researcher}\xspace}

\newcommand{\cstperson}{$\$$\textsc{Person}}
\newcommand{\cstpolitician}{$\$$\textsc{Politician}}
\newcommand{\cstpresident}{$\$$\textsc{President}}
\newcommand{\cstcountry}{$\$$\textsc{Country}}
\newcommand{\cstcity}{$\$$\textsc{City}}
\newcommand{\cststate}{$\$$\textsc{State}}
\newcommand{\cstminister}{$\$$\textsc{Minister}}
\newcommand{\cstorganization}{$\$$\textsc{Organization}}
\newcommand{\cstdigit}{$\$$\textsc{Digit}}
\newcommand{\cstdigitunit}{$\$$\textsc{DigitUnit}}
\newcommand{\cstyear}{$\$$\textsc{Year}}

\vspace{-0.1in}
\section{Introduction}
\label{sec:introduction}
Discovering \textit{textual patterns} from text data is an active research theme \cite{banko2007open,carlson2010toward,fader2011identifying,nakashole2012patty,gupta2014biperpedia}, with broad applications such as attribute extraction \cite{ghani2006text,pasca2008weakly,ravi2008using,probst2007semi}, aspect mining \cite{hu2004mining,kannan2011matching,liu2014aspect}, and slot filling \cite{yahya2014renoun,yu2016unsupervised}. Moreover, a data-driven exploration of \textit{efficient} textual pattern mining may also have strong implications on the development of efficient methods for NLP tasks on massive text corpora.

Traditional methods of textual pattern mining have made large pattern collections publicly available, but very few can extract arbitrary patterns with semantic types. Hearst patterns like ``$NP$ such as $NP$, $NP$, and $NP$'' were proposed and widely used to acquire hyponymy lexical relation \cite{hearst1992automatic}. \textsf{TextRunner} \cite{banko2007open} and \textsf{ReVerb} \cite{fader2011identifying} are blind to the typing information in their lexical patterns; \textsf{ReVerb} constrains patterns to verbs or verb phrases that end with prepositions. \textsf{NELL} \cite{carlson2010toward} learns to extract noun-phrase pairs based on a fixed set of prespecified relations with entity types like \textsf{country:president}$\rightarrow$\cstcountry$\times$\cspolitician.

\begin{figure*}
\vspace{-0.15in}
% MetaPAD generates synonymous meta-pattern groups (on "country:president") by segmentation and grouping 
% PATTY generates different SOL patterns with the shortest paths on the dependency parse trees
\subfigure[
% {Our segmentation led by pattern quality assessment considers rich contexts around the entities and determines the boundaries of the patterns when the dependency parsing does not. We study how to group synonymous meta patterns, which can produce a large collection of instances of the same relation by aggregating the extractions.}
{\textsf{MetaPAD} considers rich contexts around entities and determines pattern boundaries by pattern quality assessment while dependency parsing does not.}
]{\label{fig:comparison}\includegraphics[width=7.0in]{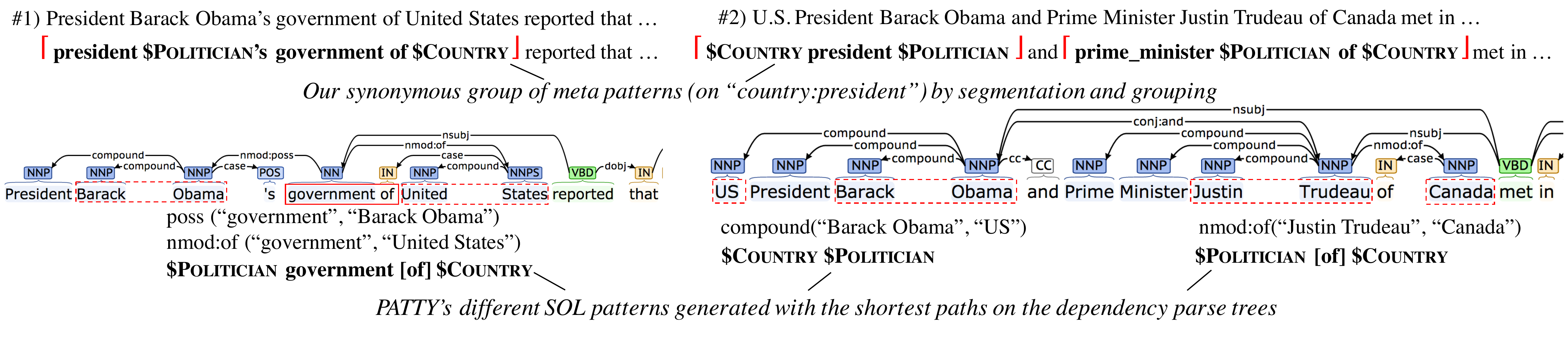}}
\vspace{-0.1in}
\subfigure[
% {Our meta pattern extends the types to both entity types and data types like \csdigit. With the pattern group, we study how to adjust the type level for appropriate granularity.}
{\textsf{MetaPAD} finds meta patterns consisting of both entity types and data types like \csdigit.  It also adjusts the type level for appropriate granularity.}
]{\label{fig:datatype}\includegraphics[width=7.0in]{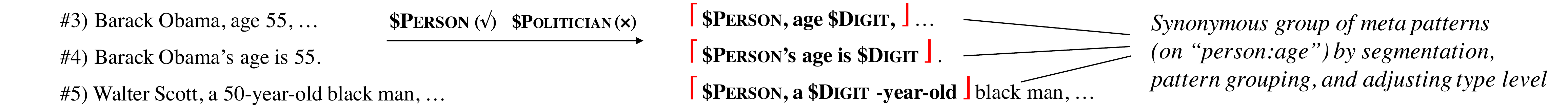}}
\vspace{-0.1in}
\caption{Comparing the synonymous group of meta patterns in \textsf{MetaPAD} with that of {SOL patterns} in \textsf{PATTY}.}
\label{fig:introduction}
\vspace{-0.15in}
\end{figure*}

One interesting exception is the SOL patterns proposed by Nakashole et al.\ in \textsf{PATTY} \cite{nakashole2012patty}. 
\textsf{PATTY} relies on the Stanford dependency parser \cite{de2006generating} and harnesses the typing information from a knowledge base \cite{auer2007dbpedia,bollacker2008freebase,nastase2010wikinet} or a typing system \cite{ling2012fine,nakashole2013fine}.
Figure~\ref{fig:comparison} shows how the SOL patterns are automatically generated with the shortest paths between two typed entities on the parse trees of individual sentences. 
Despite of the significant contributions of the work, SOL patterns have three limitations on mining typed textual patterns from a large-scale text corpus as illustrated below.
% of the  we discuss three viral features (and challenges) of mining typed textual patterns from a large-scale text corpus as follows.

First, a good typed textual pattern should be of informative, self-contained context. 
The dependency parsing in \textsf{PATTY} loses the rich context around the entities such as the word ``president'' next to ``Barack Obama'' in sentence \#1, and ``president'' and ``prime\_minister'' in \#2 (see Figure~\ref{fig:comparison}). 
Moreover, the SOL patterns are restricted to the dependency path between two entities but do not represent the data types like \csdigit for ``55'' (see Figure~\ref{fig:datatype}) and \csmonth \csday \csyear. 
Furthermore, the parsing process is costly: Its complexity is cubic in the length of sentence \cite{mcdonald2005online}, which is too costly for news and scientific corpora that often have long sentences. 
We expect an efficient textual pattern mining method for massive corpora.

Second, synonymous textual patterns are expected to be identified and grouped for handling pattern sparseness and aggregating their extractions for extending knowledge bases and question answering. 
As shown in Figure~\ref{fig:introduction}, \textsf{country:president} and \textsf{person:age} are two synonymous pattern groups.
% practitioners can easily attach attribute types like \textsf{country:president} and \textsf{person:age} to the pattern groups. 
% However, the grouping process is non-trivial. 
However, the process of finding such synonymous pattern groups is non-trivial.
Multi-faceted information should be considered: 
(1) synonymous patterns should share the same entity types or data types; 
% (2) synonymous patterns may share extractions (e.g., $\langle$United States, Barack Obama$\rangle$, $\langle$Barack Obama, 55$\rangle$) but allow different ones as well; 
(2) even for the same entity (e.g., Barack Obama), one should allow it be grouped and generalized differently (e.g., in $\langle$United States, Barack Obama$\rangle$ vs.\ $\langle$Barack Obama, 55$\rangle$); and
% (3) syonymous patterns may share the same word (e.g., ``president'') or semantically similar contextual words (e.g., ``age'' and ``-year-old''). 
(3) shared words (e.g., ``president'') or semantically similar contextual words (e.g., ``age'' and ``-year-old'') may play an important role in synonymous pattern grouping.
\textsf{PATTY} does not explore the multi-faceted information at grouping syonymous patterns, and thus cannot aggregate such extractions.

Third, the entity types in the textual patterns should be precise. In different patterns, even the same entity can be typed at different type levels. For example, the entity ``Barack Obama'' should be typed at a fine-grained level (\cstpolitician) in the patterns generated from sentence \#1--2, and it should be typed at a coarse-grained level (\cstperson) in the patterns from sentence \#3--4. However, \textsf{PATTY} does not look for appropriate granularity of the entity types.

In this paper, we propose a new typed textual pattern called \textit{meta pattern}, which is defined as follows.

\vspace{-0.05in}
\begin{definition*}[Meta Pattern]
A meta pattern refers to a frequent, informative, and precise subsequence pattern of entity types (e.g., \csperson, \cspolitician, \cscountry) or data types (e.g., \csdigit, \csmonth, \csyear), words (e.g., ``politician'', ``age'') or phrases (e.g., ``prime\_minister''), and possibly punctuation marks (e.g., ``,'', ``(''), which serves as an integral semantic unit in certain context.
\end{definition*}
\vspace{-0.05in}

We study the problem of mining meta patterns and grouping synonymous meta patterns. 
% \textit{Why do we mine the subsequence patterns from text corpus, and why do we group the synonymous meta patterns?} 
\textit{Why mining meta patterns and grouping them into synonymous meta pattern groups?}---because mining and grouping meta patterns into synonymous groups may facilitate information extraction and turning unstructured data into structures.
% Synonymous groups of meta patterns have potentials in facilitating information extraction. 
For example, given us a sentence from a news corpus, ``President Blaise Compaor$\acute{e}$'s government of Burkina Faso was founded ...'', if we have discovered the meta pattern ``president \cspolitician 's government of \cscountry'', we can recognize and type new entities (i.e., type ``Blaise Compaor$\acute{e}$'' as a \cspolitician and ``Burkina Faso'' as a \cscountry), which previously requires human expertise on language rules or heavy annotations for learning \cite{nadeau2007survey}. 
If we have grouped the pattern with synonymous patterns like ``\cscountry president \cspolitician'', we can merge the fact tuple $\langle$Burkina Faso, president, Blaise Compaor$\acute{e}$$\rangle$ into the large collection of facts of the attribute type \textsf{country:president}.

To systematically address the challenges of mining meta patterns and grouping synonymous patterns, we develop a novel framework called \textsf{MetaPAD} (\underline{Meta} \underline{PA}ttern \underline{D}iscovery). 
Instead of working on every individual sentence, our \textsf{MetaPAD} leverages \textit{massive} sentences in which redundant patterns are used to express attributes or relations of \textit{massive} instances. 
First, \textsf{MetaPAD} generates meta pattern candidates using efficient sequential pattern mining, learns a quality assessment function of the patterns candidates with a rich set of domain-independent contextual features for intuitive ideas (e.g., frequency, informativeness), and then mines the quality meta patterns by assessment-led context-aware segmentation (see Sec.~\ref{sec:segmentation}). 
Second, \textsf{MetaPAD} formulates the grouping process of synonymous meta patterns as a learning task, and solves it by integrating features from multiple facets including entity types, data types, pattern context, and extracted instances (see Sec.~\ref{sec:synonymous}). 
Third, \textsf{MetaPAD} examines the type distributions of entities in the extractions from every meta pattern group, and looks for the most appropriate type level that the patterns fit.
% to make the patterns precise.
This includes both top-down and bottom-up schemes that traverse the type ontology for the patterns' preciseness (see Sec.~\ref{sec:granularity}).

The major contributions of this paper are as follows:
(1) we propose a new definition of typed textual pattern, called \textit{meta pattern}, which is more informative, precise, and efficient in discovery than the state-of-the-art SOL pattern; (2) we develop an efficient meta-pattern mining framework, \textsf{MetaPAD}\footnote{Data and code can be found here: \url{https://github.com/mjiang89/MetaPAD}.}, of three components: generating quality meta patterns by context-aware segmentation, grouping synonymous meta patterns, and adjusting entity-type levels for appropriate granularity in the pattern groups; and (3) our experiments on three datasets of different genres---news, tweets, and biomedical corpus---demonstrate that the \textsf{MetaPAD} not only generates high quality patterns but also achieves significant improvement over the state-of-the-art in information extraction.

\section{Related Work}
\label{sec:related}
In this section, we summarize existing systems and methods that are related to the topic of this paper.

\textsf{TextRunner} \cite{banko2007open} extracts strings of words between entities in text corpus, and clusters and simplifies these word strings to produce relation-strings. 
\textsf{ReVerb} \cite{fader2011identifying} constrains patterns to verbs or verb phrases that end with prepositions. 
However, the methods in the \textsf{TextRunner}/\textsf{ReVerb} family generate patterns of frequent relational strings/phrases without entity information. 
Another line of work, open information extraction systems \cite{wu2010open,schmitz2012open,manning2014stanford,angeli2015leveraging}, are supposed to extract verbal expressions for identifying arguments. This is less related to our task of discovering textual patterns.

Google's \textsf{Biperpedia} \cite{gupta2014biperpedia,halevy2016discovering} generates \textit{$E$-$A$ patterns} (e.g., ``$A$ of $E$'' and ``$E$ 's $A$'') from users' fact-seeking queries (e.g., ``president of united states'' and ``barack oabma's wife'') by replacing entity with ``$E$'' and noun-phrase attribute with ``$A$''. \textsf{ReNoun} \cite{yahya2014renoun} generates \textit{$S$-$A$-$O$ patterns} (e.g., ``$S$'s $A$ is $O$'' and ``$O$, $A$ of $S$,'') from human-annotated corpus (e.g., ``Barack Obama's wife is Michelle Obama'' and ``Larry Page, CEO of Google'') on a pre-defined subset of the attribute names, by replacing entity/subject with ``$S$'', attribute name with ``$A$'', and value/object with ``$O$''. However, the query logs and annotations are often unavailable or expensive. Furthermore, query log word distributions are highly constrained compared with ordinary written language. So most of the $S$-$A$-$O$ patterns like ``$S$ $A$ $O$'' and ``$S$'s $A$ $O$'' will generate noisy extractions when applied to a text corpus. Textual pattern learning methods \cite{toutanova2015representing} including the above are blind to the typing information of the entities in the patterns; the patterns are not \textit{typed textual patterns}.

\textsf{NELL} \cite{carlson2010toward} learns to extract noun-phrase pairs from text corpus based on a fixed set of prespecified relations with entity types. 
\textsf{OntExt} \cite{mohamed2011discovering} clusters pattern co-occurrences for the noun-phrase pairs for a given entity type at a time and does not scale up to mining a large corpus. 
\textsf{PATTY} \cite{nakashole2012patty} was the first to harness the typing system for mining relational patterns with entity types. 
We have extensively discussed the differences between our proposed \textit{meta patterns} and \textsf{PATTY}'s \textit{SOL patterns} in the introduction: 
Meta pattern candidates are efficiently generated by sequential pattern mining \cite{agrawal1995mining,pei2004mining,zhong2012effective} on a massive corpus instead of dependency parsing on every individual sentence; 
meta pattern mining adopts a context-aware segmentation method to determine where a pattern starts and ends; and
meta patterns are not restricted to words between entity pairs but generated by pattern quality estimation based on four criteria: frequency, completeness, informativeness, and preciseness, grouped on synonymous patterns, and with type level adjusted for appropriate granularity.

\section{Meta Pattern Discovery}
\label{sec:problem}
\begin{figure}[t]
\includegraphics[width=3.4in]{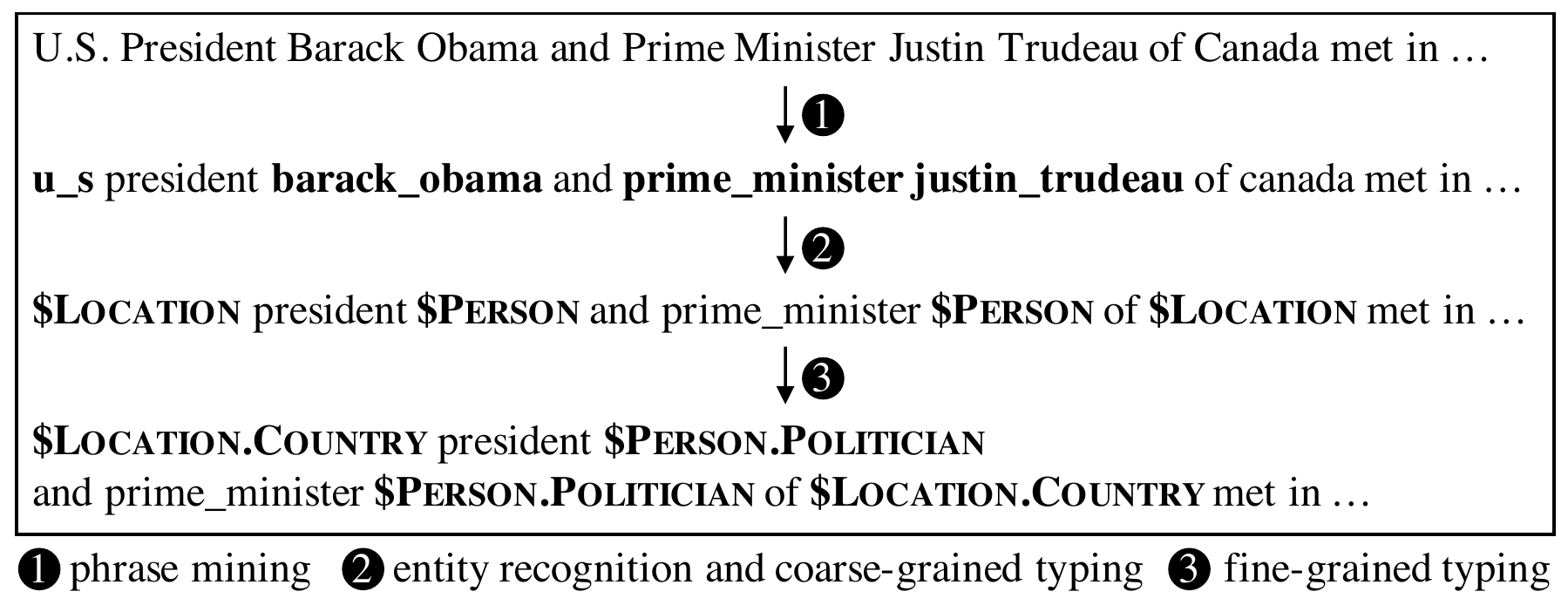}
\vspace{-0.25in}
\caption{Preprocessing for fine-grained typed corpus: given us a corpus and a typing system.}
\label{fig:preprocessing}
\vspace{-0.15in}
\end{figure}

\subsection{Preprocessing: Harnessing Typing Systems}
\label{sec:preprocessing}

To find meta patterns that are typed textual patterns, we apply efficient text mining methods for preprocessing a corpus into \textit{fine-grained typed corpus} as input in three steps as follows (see Figure~\ref{fig:preprocessing}): (1) we use a phrase mining method \cite{liu2015mining} to break down a sentence into phrases, words, and punctuation marks, which finds more real phrases (e.g., ``barack\_obama'', ``prime\_minister'') than the frequent n-grams by frequent itemset mining in \textsf{PATTY}; (2) we use a distant supervision-based method \cite{ren2015clustype} to jointly recognize entities and their coarse-grained types (i.e., \csperson, \cslocation, and \csorganization); (3) we adopt a fine-grained typing system \cite{ren2016label} to distinguish 113 entity types of 2-level ontology (e.g., \cspolitician, \cscountry, and \cscompany); we further use a set of language rules to have 6 data types (i.e., \csdigit, \csdigitunit\footnote{\csdigitunit: ``percent'', ``\%'', ``hundred'', ``thousand'', ``million'', ``billion'', ``trillion''...}, \csdigitrank\footnote{\csdigitrank: ``first'', ``1st'', ``second'', ``2nd'', ``44th''...}, \csmonth, \csday, and \csyear). 
Now we have a fine-grained, typed corpus consisting of the same kinds of tokens as defined in the meta pattern: entity types, data types, phrases, words, and punctuation marks. All the tools are publicly available on \textsf{GitHub}.

\subsection{The Proposed Problem}
\label{sec:theproblem}

\begin{problem*}[Meta Pattern Discovery]
Given a fine-grained, typed corpus of massive sentences $\mathcal{C} = [\ldots, S, \ldots]$, and each sentence is denoted as $S = t_1t_2\ldots t_n$ in which $t_k \in \mathcal{T}\cup\mathcal{P}\cup\mathcal{M}$ is the $k$-th token ($\mathcal{T}$ is the set of entity types and data types, $\mathcal{P}$ is the set of phrases and words, and $\mathcal{M}$ is the set of punctuation marks), the task is to find \textbf{synonymous groups of quality meta patterns}. 
A \emph{meta pattern} $mp$ is a subsequential pattern of the tokens from the set $\mathcal{T}\cup\mathcal{P}\cup\mathcal{M}$. 
A \emph{synonymous meta pattern group} is denoted by $\mathcal{MPG} = [\ldots, mp_i, \ldots, mp_j \ldots]$ in which each pair of meta patterns, $mp_i$ and $mp_j$, are synonymous.
\end{problem*}

\textit{What is a quality meta pattern?} 
Here we take the sentences as sequences of tokens. 
Previous sequential pattern mining algorithms mine frequent subsequences satisfying a single metric, the minimum support threshold (\textsf{min\_sup}), in a transactional sequence database \cite{agrawal1995mining}.
However, for text sequence data, the quality of our proposed textual pattern, the meta pattern, should be evaluated similar to phrase mining \cite{liu2015mining}, in four criteria as illustrated below.
 % here are examples of comparisons between meta patterns of high quality and low quality.
\begin{example*}
The quality of a pattern is evaluated with the following criteria: (the former pattern has higher quality than the latter)
\item \textit{Frequency:} ``\csdigitrank president of \cscountry'' vs.\ ``young president of \cscountry'';
\item \textit{Completeness:} ``\cscountry president \cspolitician'' vs.\ ``\cscountry president'', ``\csperson's wife, \csperson'' vs.\ ``\csperson's wife'';
\item \textit{Informativeness:} ``\csperson's wife, \csperson'' vs.\ ``\csperson and \csperson'';
\item \textit{Preciseness:} ``\cscountry president \cspolitician'' vs.\ ``\cslocation president \csperson'', ``\csperson's wife, \csperson'' vs.\ ``\cspolitician's wife, \csperson'', ``population of \cslocation'' vs.\ ``population of \cscountry''.
\end{example*}

\textit{What are synonymous meta patterns?} 
The full set of frequent sequential patterns from a transaction dataset is huge \cite{agrawal1995mining}; 
and the number of meta patterns from a massive corpus is also big. 
Since there are multiple ways to express the same or similar meanings in a natural language, many meta patterns may share the same or nearly the same meaning. 
Examples have been given in Figure~\ref{fig:introduction}.
Grouping synonymous meta patterns can help aggregate a large number of extractions of different patterns from different sentences. 
And the type distribution of the aggregated extractions can help us adjust the meta patterns in the group for preciseness.

\section{The MetaPAD Framework}
\label{sec:approach}
\begin{figure}
\includegraphics[width=3.4in]{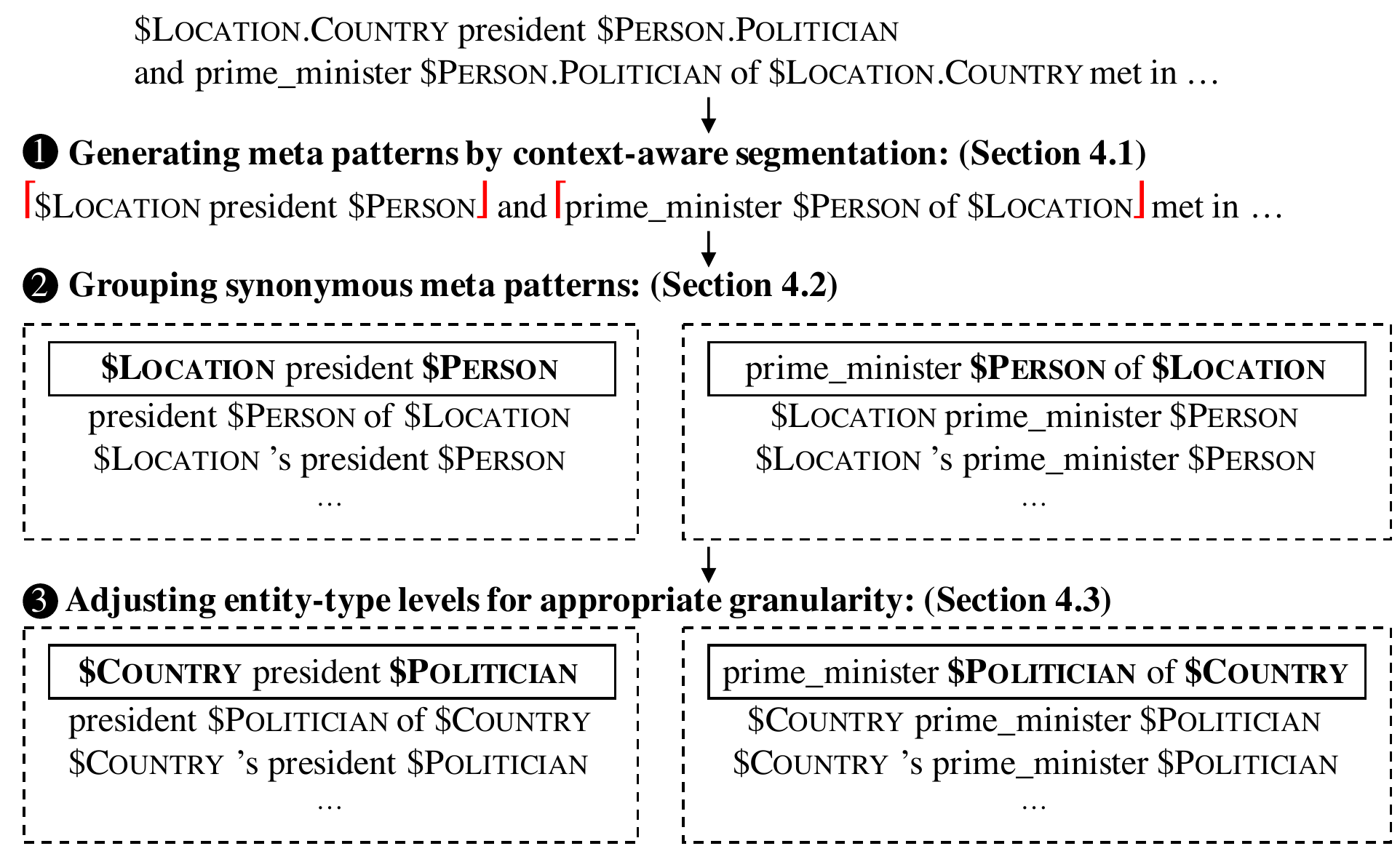}
\vspace{-0.25in}
\caption{Three modules in our \textsf{MetaPAD} framework.}
\label{fig:framework}
\vspace{-0.2in}
\end{figure}

Figure~\ref{fig:framework} presents the \textsf{MetaPAD} framework for \underline{Meta} \underline{PA}ttern \underline{D}iscovery. It has three modules. First, it develops a context-aware segmentation method to determine the boundaries of the subsequences and generate the meta patterns of frequency, completeness, and informativeness (see Sec.~\ref{sec:segmentation}). Second, it groups synonymous meta patterns into clusters (see Sec.~\ref{sec:synonymous}). Third, for every synonymous pattern group, it adjusts the levels of entity types for appropriate granularity to have precise meta patterns (see Sec.~\ref{sec:granularity}).

\subsection{Generating meta patterns by context-aware segmentation}
\label{sec:segmentation}

\paragraph{Pattern candidate generation.} We adopt the standard frequent sequential pattern mining algorithm \cite{pei2004mining} to look for pattern candidates that satisfy a \textsf{min\_sup} threshold.
In practice, one can set a maximum pattern length $\omega$ to restrict the number of tokens in the patterns. Different from syntactic analysis of very long sentences, our meta pattern mining explores pattern structures that are local but still of wide context: in our experiments, we set $\omega = 20$.

\paragraph{Meta pattern quality assessment.} Given a huge number of pattern candidates that can be messy (e.g., ``of \cscountry'' and ``\cspolitician and''), it is desired but challenging to assess the quality of the patterns with a very few training labels. 
We introduce a rich set of \textit{contextual features} of the patterns according to the quality criteria (see Sec.~\ref{sec:theproblem}) as follows and train a classifier to estimate the quality function $Q(mp) \in [0, 1]$ where $mp$ is a meta pattern candidate:

\noindent \textbf{1. Frequency:} A good pattern $mp$ should occur with sufficient count $c(mp)$ in a given typed text corpus.

\noindent \textbf{2. Concordance:} If the collocation of tokens in such frequency that is significantly higher than what is expected due to chance, the meta pattern $mp$ has good concordance. To statistically reason about the concordance, we consider a null hypothesis: the corpus is generated from a series of independent Bernoulli trials. Suppose the number of tokens in the corpus is $L$ that can be assumed to be fairly large. The expected frequency of a pair of sub-patterns $\langle$${mp}_{l}$,${mp}_{r}$$\rangle$ under our null hypothesis of their independence is
\begin{equation}
{\mu}_0(c(\langle {mp}_{l}, {mp}_{r} \rangle)) = L \cdot p({mp}_{l}) \cdot p({mp}_{r}),
\end{equation}
where $p(mp) = \frac{c(mp)}{L}$ is the empirical probability of the pattern.
We use Z score to provide a quantitative measure of a pair of sub-patterns $\langle$${mp}_{l}$,${mp}_{r}$$\rangle$ forming the best collocation as $mp$ in the corpus:
\begin{equation}
Z(mp) = \max_{\langle {mp}_{l}, {mp}_{r} \rangle = mp} \frac{c(mp) - {\mu}_0(c(\langle {mp}_{l}, {mp}_{r} \rangle))}{{\sigma}_{\langle {mp}_{l}, {mp}_{r} \rangle}},
\end{equation}
where ${\sigma}_{\langle {mp}_{l}, {mp}_{r} \rangle}$ is the standard deviation of the frequency. A high Z score indicates that the pattern is acting as an integral semantic unit in the context: its composed sub-patterns are highly associated.

\noindent \textbf{3. Informativeness:} A good pattern $mp$ should have informative context. We examine the counts of different kinds of tokens (e.g., types, words, phrases, non-stop words, marks). For example, the pattern ``\csperson 's wife \csperson'' is informative for the non-stop word ``wife''; ``\csperson was born\_in \cscity'' is good for the phrase ``born\_in''; and ``\csperson, \csdigit,'' is also informative for the two different types and two commas.

\noindent \textbf{4. Completeness:} We use the ratio between the frequencies of the pattern candidate (e.g., ``\cscountry president \cspolitician'') and its sub-patterns (e.g., ``\cscountry president''). If the ratio is high, the candidate is likely to be complete. We also use the ratio between the frequencies of the pattern candidate and its super-patterns. If the ratio is high, the candidate is likely to be incomplete. Moreover, we expect the meta pattern to be NOT bounded by stop words. For example, neither ``and \cscountry president'' nor ``president \cspolitician and'' is properly bounded.

\noindent \textbf{5. Coverage:} A good typed pattern can extract multiple instances. For example, the type \cspolitician in the pattern ``\cspolitician's healthcare law'' refers to only one entity ``Barack Obama'', and thus has too low coverage in the corpus.

\begin{figure}
\includegraphics[width=3.4in]{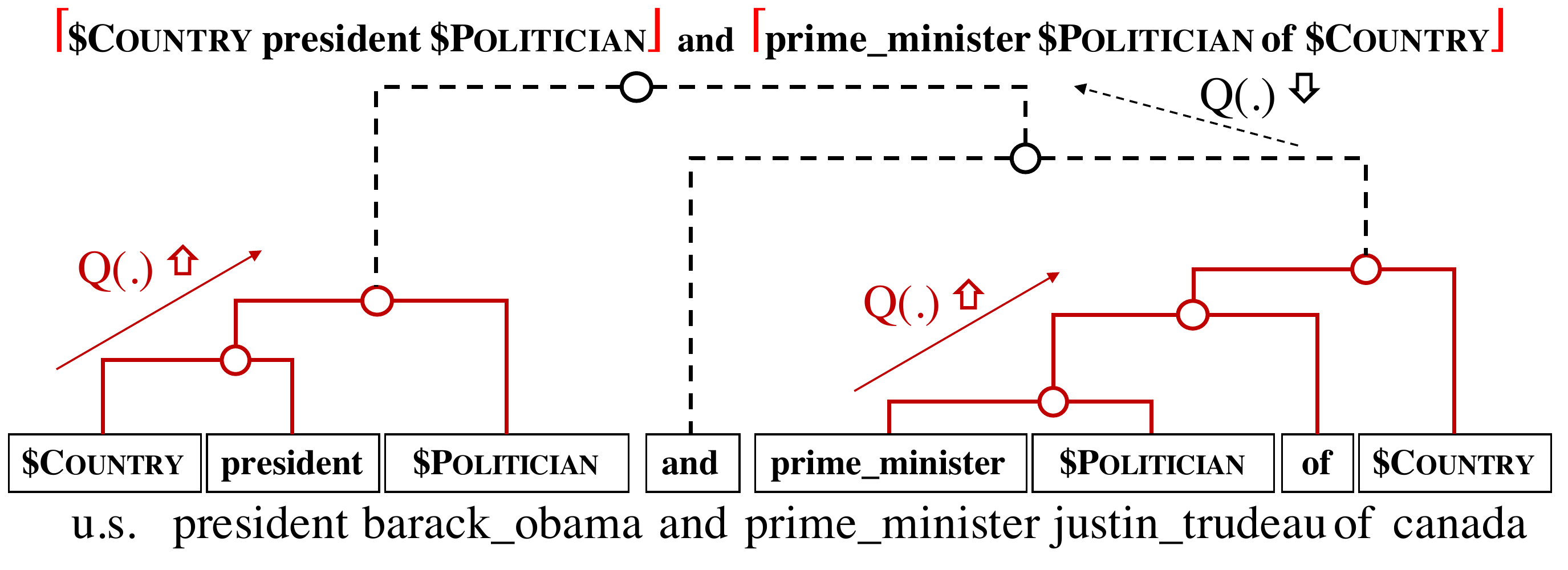}
\vspace{-0.3in}
\caption{Generating meta patterns by context-aware segmentation with the pattern quality function $Q(.)$.}
\label{fig:segmentation}
\vspace{-0.2in}
\end{figure}

We train a classifier based on random forests \cite{breiman2001random} for learning the meta-pattern quality function $Q(mp)$ with the above rich set of contextual features. Our experiments (not reported here for the sake of space) show that using only 100 pattern labels can achieve similar precision and recall as using 300 labels. Note that the learning results can be transferred to other domains: the features of low-quality patterns ``\cspolitician and \cscountry'' and ``\csbacteria and \csantibiotics'' are similar; the features of high-quality patterns ``\cspolitician is president of \cscountry'' and ``\csbacteria is resistant to \csantibiotics'' are similar.

\begin{table*}[t]
\vspace{-0.10in}
\caption{Issues of quality over-/under-estimation can be fixed when the segmentation rectifies pattern frequency.}
\vspace{-0.15in}
\label{tab:rectified}
\begin{tabular}{l|rr|rrr}
\toprule
& \multicolumn{2}{|c|}{Before segmentation} & \multicolumn{3}{|c}{Frequency rectified after segmentation} \\
Pattern candidate & Count & Quality & Count & Quality & Issue fixed by feedback \\ \hline
\cscountry president \cspolitician & 2,912 & 0.93 & 2,785 & 0.97 & N/A \\
prime\_minister \cspolitician of \cscountry & 1,285 & 0.84 & 1,223 & 0.92 & slight underestimation \\
\cspolitician and prime\_minister \cspolitician & 532 & 0.70 & 94 & 0.23 & overestimation \\
\bottomrule
\end{tabular}
\vspace{-0.1in}
\end{table*}

\paragraph{Context-aware segmentation using $Q(.)$ with feedback.} With the pattern quality function $Q(.)$ learnt from the rich set of contextual features, we develop a bottom-up segmentation algorithm to construct the best partition of segments of high quality scores. As shown in Figure~\ref{fig:segmentation}, we use $Q(.)$ to determine the boundaries of the segments: we take  ``\cscountry president \cspolitician'' for its high quality score; we do not take the candidate ``and prime\_minister \cspolitician of \cscountry'' because of its low quality score.

Since $Q(mp)$ was learnt with features including the raw frequency $c(mp)$, the quality score may be overestimated or underestimated: the principle is that every token's occurrence should be assigned to only one pattern but the raw frequency may count the tokens multiple times. Fortunately, after the segmentation, we can rectify the frequency as $c_{r}(mp)$, for example in Figure~\ref{fig:segmentation}, the segmentation avoids counting ``\cspolitician and prime\_minister \cspolitician'' of overestimated frequency/quality (see Table~\ref{tab:rectified}).

Once the frequency feature is rectified, we re-learn the quality function $Q(.)$ using $c(mp)$ as feedback and re-segment the corpus with it. This can be an iterative process but we found in only one iteration, the result converges. Algorithm~\ref{alg:segmentation} shows the details.

\begin{algorithm}[h]
\caption{Context-aware segmentation using $Q$ with feedback}
\begin{algorithmic}[1]
\Require corpus of sentences $\mathcal{C}$=[..., $S$, ...], $S = t_1t_2...t_n$ ($t_k$ is the $k$-th token), a set of meta pattern candidates $MP_{cand}$, meta-pattern quality function $Q(.)$ learnt by contextual features
\State Set all the rectified frequency $c_{r}(mp)$ to zero
\For{$S \in \mathcal{C}$}
	\State Segment the sentence $S$ into $Seg$=[..., $mp$, ...] by maximizing $\sum_{mp \in Seg}{Q(mp)}$ with a bottom-up scheme (see Figure~\ref{fig:segmentation}), where $mp \in MP_{cand}$ is a segment of high quality score
	\For{$mp \in Seg$}
		\State $c_{r}(mp) \leftarrow c_{r}(mp) + 1$
	\EndFor
\EndFor
\State Re-learn $Q(.)$ by replacing the raw frequency feature $c(mp)$ with the rectified frequency $c_r(mp)$ as feedback
\State Re-segment the corpus $\mathcal{C}$ with the new $Q(.)$ \\
\Return{Segmented corpus, a set of quality meta patterns in the segmented corpus, and their quality scores in $Q(.)$}
\end{algorithmic}
\label{alg:segmentation}
\end{algorithm}

%\vspace{-0.2in}
\subsection{Grouping synonymous meta patterns}
\label{sec:synonymous}

\begin{figure}
\includegraphics[width=3.4in]{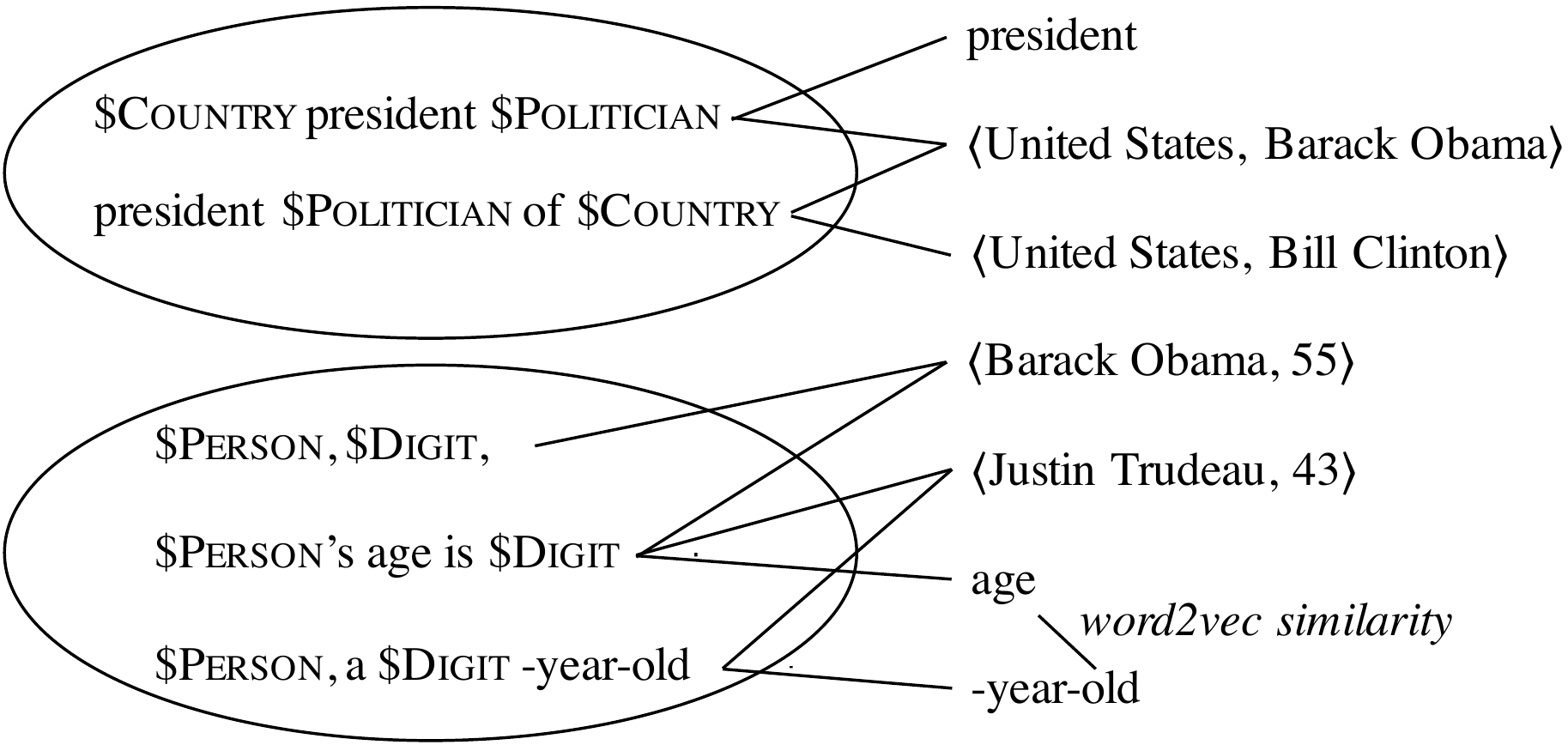}
\vspace{-0.25in}
\caption{Grouping synonymous meta patterns with information of context words and extractions.}
\label{fig:synonymous}
\vspace{-0.15in}
\end{figure}

Grouping truly synonymous meta patterns enables a large collection of extractions of the same relation aggregated from different but synonymous patterns. For example, there could be hundreds of ways of expressing the relation \textsf{country:president}; if we group all such meta patterns, we can aggregate all the extractions of this relation from massive corpus. \textsf{PATTY} \cite{nakashole2012patty} has a narrow definition of their synonymous dependency path-based SOL patterns: two patterns are synonymous if they generate the same set of extractions from the corpus. Here we develop a learning method to incorporate information of three aspects, (1) entity/data types in the pattern, (2) context words/phrases in the pattern, and (3) extractions from the pattern, to assign the meta patterns into groups. Our method is based on three assumptions as follows (see Figure~\ref{fig:synonymous}):

\noindent \textbf{\textit{A1:}} Synonymous meta patterns must have the same entity/data types: the meta patterns ``\csperson's age is \csdigit'' and ``\csperson's wife is \csperson'' cannot be synonymous;

\noindent \textbf{\textit{A2:}} If two meta patterns share (nearly) the same context words/phrases, they are more likely to be synonymous: the patterns ``\cscountry president \cspolitician'' and ``president \cspolitician of \cscountry'' share the word ``president'';

\noindent \textbf{\textit{A3:}} If two patterns generate more common extractions, they are more likely to be synonymous: both ``\csperson's age is \csdigit'' and ``\csperson, \csdigit,'' generate $\langle$Barack Obama, 55$\rangle$.

Since the number of groups cannot be pre-specified, we propose to first construct a pattern-pattern graph in which the two pattern nodes of every edge satisfy \textit{A1} and are predicted to be synonymous, and then use a clique detection technique \cite{harary1957procedure} to find all the cliques as synonymous meta patten groups. Each pair of the patterns ($mp_i$, $mp_j$) in the group $\mathcal{MPG} = [\ldots, mp_i, \ldots, mp_j \ldots]$ are synonymous.

For the graph construction, we train Support Vector Regression Machines \cite{drucker1997support} to learn the following features of a pair of patterns based on \textit{A2} and \textit{A3}: (1) the numbers of words, non-stop words, phrases that each pattern has and they share; (2) the maximum similarity score between pairs of non-stop words or phrases in the two patterns; (3) the number of extractions that each pattern has and they share. The similarity between words/phrases is represented by the cosine similarity of their word2vec embeddings \cite{mikolov2013distributed,toutanova2015representing}.

\subsection{Adjusting type levels for preciseness}
\label{sec:granularity}

\begin{figure}
\includegraphics[width=3.4in]{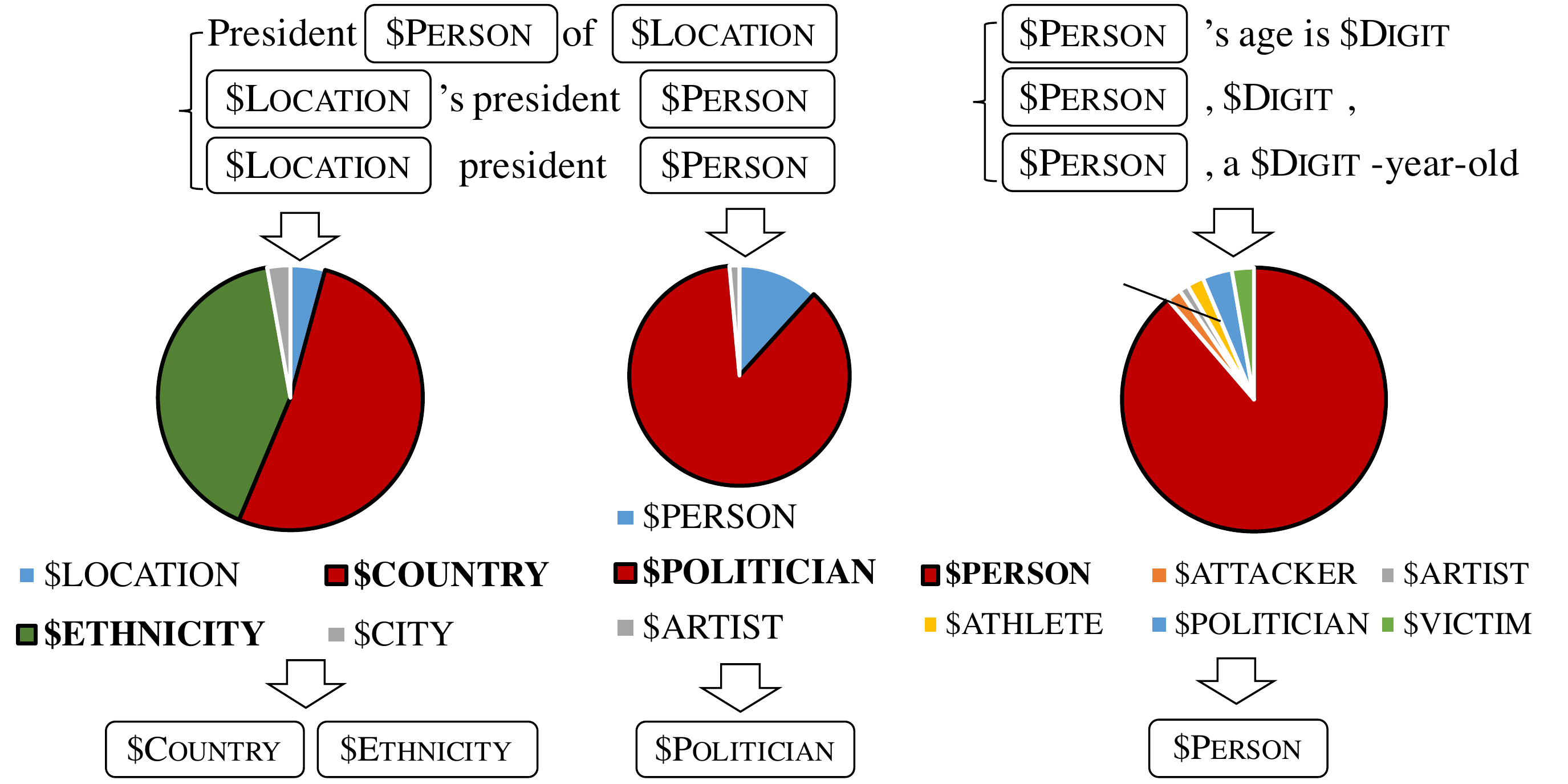}
\vspace{-0.25in}
\caption{Adjusting entity-type levels for appropriate granularity with entity-type distributions.}
\label{fig:granularity}
\vspace{-0.2in}
\end{figure}

Given a group of synonymous meta patterns, we expect the patterns to be precise: it is desired to determine the levels of the entity types in the patterns for appropriate granularity. Thanks to the grouping process of synonymous meta patterns, we have rich type distributions of the entities from the large collection of extractions.

As shown in Figure~\ref{fig:granularity}, given the ontology of entity types (e.g., \cslocation: \cscountry, \csstate, \cscity, \ldots; \csperson: \csartist, \csathlete, \cspolitician, \ldots), for the group of synonymous meta patterns ``president \csperson of \cslocation'', ``\cslocation's president \csperson'', and ``\cslocation president \csperson'', are the entity types, \cslocation and \csperson, of appropriate granularity to make the patterns precise? If we look at the type distributions of entities in the extractions of these patterns, it is clear that most of the entities for \cslocation are typed at a fine-grained level as \cscountry (e.g., ``United States'') or \csethnicity (e.g., ``Russian''), and most of the entities for \csperson also have the fine-grained type \cspolitician. Therefore, compared with ``\cslocation president \csperson'', the two fine-grained meta patterns ``\cscountry president \cspolitician'' and ``\csethnicity president \cspolitician'' are more precise; we have the same claim for other meta patterns in the synonymous group. On the other hand, for the group of synonymous meta patterns on \textsf{person:age}, we can see most of the entities are typed at a coarse-grained level as \csperson instead of \csathlete or \cspolitician. So the entity type in the patterns is good to be \csperson. From this observation, given an entity type $T$ in the meta pattern group, we propose a metric, called \textit{graininess}, that is defined as the fraction of the entities typed by $T$ that can be fine-grained to $T$'s sub-types:
\begin{equation}
g(T) = \frac{{\sum}_{T' \in subtype\_of(T)}{num\_entity(T')}}{{\sum}_{T' \in subtype\_of(T) \cup \{T\}}{num\_entity(T')}}.
\end{equation}
If $g(T)$ is higher than a threshold $\theta$, we go down the type ontology for the fine-grained types.

Suppose we have determined the appropriate type level in the meta pattern group using the graininess metric. However, not every type at the level should be used to construct precise meta patterns. For example, we can see from Figure~\ref{fig:granularity} for the patterns on \textsf{president}, very few entities of \cslocation are typed as \cscity, and very few entities of \csperson are typed as \csartist. Comparing with \cscountry, \csethnicity, and \cspolitician, these fine-grained types are at the same level but have too small support of extractions. We exclude them from the meta pattern group. Based on this idea, for an entity type $T$, we propose another metric, called \textit{support}, that is defined as the ratio of the number of entities typed by $T$ to the maximum number of entities typed by $T$'s sibling types:
\begin{equation}
s(T) = \frac{num\_entity(T)}{{\max}_{T' \in sibling-type\_of(T) \cup \{T\}}{num\_entity(T')}}.
\end{equation}
If $s(T)$ is higher than a threshold $\gamma$, we consider the type $T$ in the meta pattern group; otherwise, we drop it.

With these two metrics, we develop a \textit{top-down} scheme that first conducts segmentation and synonymous pattern grouping on the coarse-grained typed meta patterns, and then checks if the fine-grained types are significant and if the patterns can be split to the fine-grained level; we also develop a \textit{bottom-down} scheme that first works on the fine-grained typed meta patterns, and then checks if the patterns can be merged into a coarse-grained level.

\subsection{Complexity analysis}
\label{sec:complexity}

We develop three new components in our \textsf{MetaPAD}. The time complexity of generating meta patterns with context-aware segmentation is $\mathcal{O}(\omega |\mathcal{C}|)$ where $\omega$ is the maximum pattern length and $|\mathcal{C}|$ is the corpus size (i.e., the total number of tokens in the corpus). The complexity of grouping synonymous meta patterns is $\mathcal{O}(|\mathcal{MP}|)$, and the complexity of adjusting type levels is $\mathcal{O}(h |\mathcal{MP}|)$ where $|\mathcal{MP}|$ is the number of quality meta patterns and $h$ is the height of type ontology. The total complexity is $\mathcal{O}(\omega |\mathcal{C}|+(h+1) |\mathcal{MP}|)$, which is linear in the corpus size.

\textsf{PATTY} \cite{nakashole2012patty} is also scalable in the number of sentences but for each sentence, the complexity of dependency parsing it adopted is as high as $\mathcal{O}(n^3)$ where $n$ is the length of the sentence. If the corpus has many long sentences, \textsf{PATTY} is time-consuming; whereas our \textsf{MetaPAD}'s complexity is linear to the sentence length for every individual sentence. The empirical study on the scalability can be found in the next section.

\section{Experiments}
\label{sec:experiments}
This section reports our essential experiments that demonstrate the effectiveness of the \textsf{MetaPAD} at (1) typed textual pattern mining: discovering synonymous groups of meta patterns, and (2) one application: extracting tuple information from three datasets of different genres. Additional results regarding efficiency are reported as well.

\begin{table}
\caption{Three datasets of different genres.}
\vspace{-0.15in}
\label{tab:dataset}
\Scale[0.95]{\begin{tabular}{lrrrr}
\toprule
Dataset & File Size & \#Document & \#Entity & \#Entity Mention \\
\hline
\textsf{APR} (news) & 199MB & 62,146 & 284,061 & 6,732,399 \\
\textsf{TWT} (tweet) & 1.05GB & 13,200,821 & 618,459 & 21,412,381 \\
\textsf{CVD} (paper) & 424MB & 463,040 & 751,158 & 27,269,242 \\
\bottomrule
\end{tabular}}
\vspace{-0.1in}
\end{table}

\subsection{Datasets}

\leftmargini=12pt

Table~\ref{tab:dataset} presents the statistics of three datasets from different genres:
% \begin{compactitem}
\begin{itemize}
\parskip -0.2ex
\item \textsf{APR}: news from The \underline{A}ssociated \underline{P}ress and \underline{R}euters in 2015;
\item \textsf{TWT}: tweets collected via \underline{Tw}i\underline{t}ter API in 2015/06--2015/09;
\item \textsf{CVD}: paper titles and abstracts about the \underline{C}ardio\underline{v}ascular \underline{d}iseases from the \textsf{PubMed} database.
\end{itemize}
% \end{compactitem}
The news and biomedical paper corpora often have long sentences, which is rather challenging for textual pattern mining. For example, the component of dependency parsing in \textsf{PATTY} \cite{nakashole2012patty} has cubic computational complexity of the length for individual sentences.

The preprocessing techniques in our \textsf{MetaPAD} adopt distant supervision with external databases for entity recognition and fine-grained typing (see Sec.~\ref{sec:preprocessing}). For the general corpora like news and tweets, we use \textsf{DBpedia} \cite{auer2007dbpedia} and \textsf{Freebase} \cite{bollacker2008freebase}; for the biomedical corpus, we use public \textsf{MeSH} databases \cite{mesh}.

\subsection{Experimental Settings}

\begin{table}
\caption{Entity-Attribute-Value tuples as ground truth.}
\vspace{-0.15in}
\label{tab:truth}
\Scale[0.95]{\begin{tabular}{lllr}
\toprule
Attribute & Type of Entity & Type of Value & \#Tuple \\
\hline
\textsf{country:president} & \cscountry & \cspolitician & 1,170 \\
\textsf{country:minister} & \cscountry & \cspolitician & 1,047 \\
\textsf{state:representative} & \csstate & \cspolitician & 655 \\
\textsf{state:senator} & \csstate & \cspolitician & 610 \\
\textsf{county:sheriff} & \cscounty & \cspolitician & 106 \\
\textsf{company:ceo} & \cscompany & \csbusinessperson & 1,052 \\
\textsf{university:professor} & \csuniversity & \csresearcher & 707 \\
\textsf{award:winner} & \csaward & \csperson & 274 \\
\bottomrule
\end{tabular}}
\vspace{-0.15in}
\end{table}

\begin{table*}
\vspace{-0.1in}
\caption{Synonymous meta patterns and their extractions that \textsf{MetaPAD} generates from the \textit{biomedical corpus} \textsf{CVD}.}
\vspace{-0.15in}
\label{tab:discoverycvd}
\Scale[0.7]{\begin{tabular}{l||ll}
\toprule
{A \textbf{group} of synonymous meta patterns} & {\cstreatment} & {\csdisease} \\ \hline
{\cstreatment was used to treat \csdisease} & {zoledronic acid therapy} & {Paget's disease of bone} \\
{\csdisease using the \cstreatment} & {bisphosphonates} & {osteoporosis} \\
{\cstreatment has been widely used to treat \csdisease} & {calcitonin} & {Paget's disease of bone} \\
{\cstreatment of patients with \csdisease} & {calcitonin} & {osteoporosis} \\
{...} & {...} & {...} \\
\bottomrule
\toprule
{A \textbf{group} of synonymous meta patterns} & {\csbacteria} & {\csantibiotics} \\ \hline
{\csbacteria was resistant to \csantibiotics} & {corynebacterium striatum BM4687} & {gentamicin} \\
{\csbacteria are resistant to \csantibiotics} & {corynebacterium striatum BM4687} & {tobramycin} \\
{\csbacteria is the most resistant to \csantibiotics} & {methicillin-susceptible S aureus} & {vancomycin} \\
{\csbacteria, particularly those resistant to \csantibiotics} & {multidrug-resistant enterobacteriaceae} & {gentamicin} \\
{...} & {...} & {...} \\
\bottomrule
\end{tabular}}
\vspace{-0.05in}
\end{table*}

\begin{table}
\vspace{-0.1in}
\caption{Synonymous meta patterns and their extractions that \textsf{MetaPAD} generates from the \textit{news corpus} \textsf{APR} on \textsf{country:president}, \textsf{company:ceo}, and \textsf{person:date\_of\_birth}.}
\vspace{-0.15in}
\label{tab:discoverynews}
\Scale[0.62]{\begin{tabular}{l||ll}
\toprule
{A \textbf{group} of synonymous meta patterns} & {\cscountry} & {\cspolitician} \\ \hline
{\cscountry president \cspolitician} & {United States} & {Barack Obama} \\
{\cstcountry's president \cspolitician} & {United States} & {Bill Clinton} \\
{president \cspolitician of \cscountry} & {Russia} & {Vladimir Putin} \\
{\cspolitician, the president of \cscountry,} & {France} & {Fran\c{c}ois Hollande} \\
{...} & {...} & {...} \\
{president \cspolitician's government of \cscountry} & {Comoros} & {Ikililou Dhoinine} \\
{\cspolitician was elected as the president of \cscountry} & {Burkina Faso} & {Blaise Compaor$\acute{e}$} \\
\bottomrule
\toprule
{A \textbf{group} of synonymous meta patterns} & {\cscompany} & {\csbusinessperson} \\ \hline
{\cscompany ceo \csbusinessperson} & {Apple} & {Tim Cook} \\
{\cscompany chief executive \csbusinessperson} & {Facebook} & {Mark Zuckerburg} \\
{\csbusinessperson, the \cscompany ceo,} & {Hewlett-Packard} & {Carly Fiorina} \\
{\cscompany former ceo \csbusinessperson} & {Yahoo!} & {Marissa Mayer} \\
{...} & {...} & {...} \\
{\csbusinessperson was appointed as ceo of \cscompany} & {Infor} & {Charles Phillips} \\
{\csbusinessperson, former interim ceo, leaves \cscompany} & {Afghan Citadel} & {Roya Mahboob} \\
\bottomrule
\toprule
{A \textbf{group} of synonymous meta patterns} & \multicolumn{2}{l}{{\csperson} \qquad\qquad\qquad {\csday \csmonth \csyear}} \\ \hline
{\csperson was born \csmonth \csday, \csyear} & {Willie Howard Mays} & {6 May 1931} \\
{\csperson was born on \csday \csmonth \csyear} &  {Robert David Simon} & {29 May 1941} \\
{\csperson (born on \csmonth \csday, \csyear)} & {Phillip Joel Hughes} & {30 Nov 1988} \\
{\csperson (born on \csday \csmonth \csyear)} & {...} & {...} \\
{\csperson, was born on \csmonth \csday, \csyear} & {Carl Sessions Stepp} & {8 Sept 1956} \\
{...} & {Richard von Weizsaecker} & {15 April 1920} \\
\bottomrule
\end{tabular}}
\vspace{-0.2in}
\end{table}

We conduct two tasks in the experiments. The first task is to \textit{discover typed textual patterns from massive corpora and organize the patterns into synonymous groups}. We compare with the state-of-the-art SOL pattern synset mining method \textbf{\textsf{PATTY}} \cite{nakashole2012patty} on both the quality of patterns and the quality of synonymous pattern groups. Since there is no standard ground truth of the typed textual patterns, we report extensive qualitative analysis on the three datasets.

The second task is to \textit{extract $\langle$\underline{e}ntity, \underline{a}ttribute, \underline{v}alue$\rangle$ $($EAV$)$ tuple information}. For every synonymous pattern set generated by the competitive methods from news and tweets, we assign it to one attribute type from the set in Table~\ref{tab:truth} if appropriate. We collect 5,621 EAV-tuples from the extractions, label them as true or false, and finally, we have 3,345 true EAV-tuples. We have 2,400 true EAV-tuples from \textsf{APR} and 2,090 from \textsf{TWT}. Most of them are out of the existing knowledge bases: we are exploring new extractions from new text corpora.

We evaluate the performance in terms of \textit{precision} and \textit{recall}. Precision is defined as the fraction of the predicted EAV-tuples that are true. Recall is defined as the fraction of the labelled true EAV-tuples that are predicted as true EAV-tuples. We use (1) the F1 score that is the harmonic mean of precision and recall, and (2) the Area Under the precision-recall Curve (AUC). All the values are between 0 and 1, and a higher value means better performance.

In the second task, besides \textbf{\textsf{PATTY}}, the competitive methods for tuple extraction are: \textbf{\textsf{Ollie}} \cite{schmitz2012open} is an open IE system that extracts relational tuples with syntactic and lexical patterns; \textbf{\textsf{ReNoun}} \cite{yahya2014renoun} learns ``$S$-$A$-$O$'' patterns such as ``$S$ $A$, $O$,'' and ``$A$ of $S$ is $O$'' with annotated corpus. Both methods ignore the entity-typing information.
We develop four alternatives of \textsf{MetaPAD} as follows:

\noindent \textbf{1. \textsf{MetaPAD-T}} only develops segmentation to generate patterns in which the entity types are at the \textit{\underline{t}op} (coarse-grained) level;

\noindent \textbf{2. \textsf{MetaPAD-TS}} develops all the three components of \textsf{MetaPAD} including \underline{s}ynonymous pattern grouping based on \textsf{MetaPAD-T};

\noindent \textbf{3. \textsf{MetaPAD-B}} only develops segmentation to generate patterns in which the entity types are at the \textit{\underline{b}ottom} (fine-grained) level;

\noindent \textbf{4. \textsf{MetaPAD-BS}} develops all the three components of \textsf{MetaPAD} including \underline{s}ynonymous pattern grouping based on \textsf{MetaPAD-B}.

For the parameters in \textsf{MetaPAD}, we set the maximum pattern length as $\omega = 20$, the threshold of graininess score as $\theta = 0.8$, and the threshold of support score as $\gamma = 0.1$.

\subsection{Results on Typed Textual Pattern Discovery}

\begin{figure}
\vspace{-0.1in}
\includegraphics[width=3.4in]{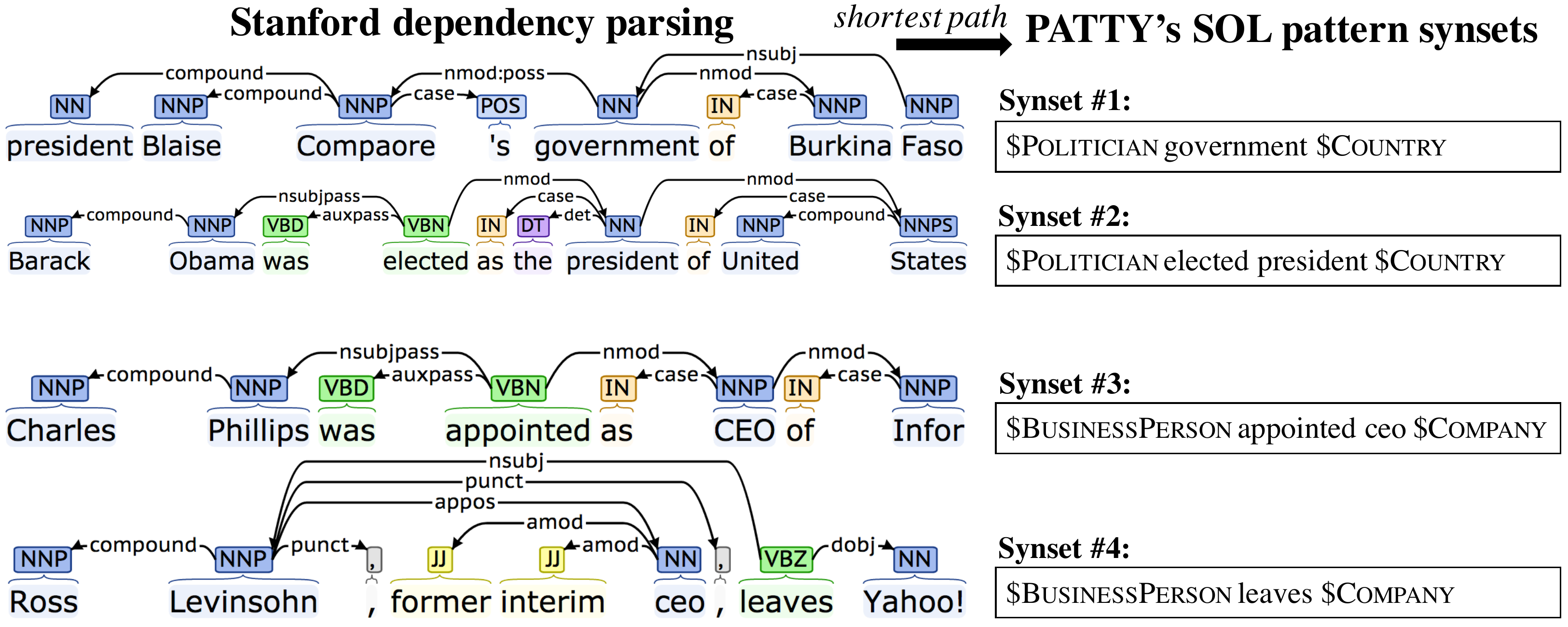}
\vspace{-0.24in}
\caption{Compared with our meta patterns, the SOL pattern mining does not take the rich context into full consideration of pattern quality assessment; the definition of SOL pattern synset is too limited to group truly synonymous patterns.}
\label{fig:patty}
\vspace{-0.2in}
\end{figure}

Our proposed \textsf{MetaPAD} discovers high-quality meta patterns by context-aware segmentation from massive text corpus with a pattern quality assessment function. It further organizes them into synonymous groups. With each group of the truly synonymous meta patterns, we can easily assign an appropriate attribute type to it, and harvest a large collection of instances extracted by different patterns of the same group. All these can be done not only on the news corpus but also on the biomedical corpus.

Table~\ref{tab:discoverynews} presents the groups of synonymous meta patterns that express attribute types \textsf{country:president} and \textsf{company:ceo}. 
\textbf{First}, we can see that the meta patterns are generated from a typed corpus instead of the shortest path of a dependency parse tree. 
Thus, the patterns can keep rich, wide context information. 
\textbf{Second}, the meta patterns are of high quality on informativeness, completeness, and so on, and practitioners can easily tell why the patterns are extracted as an integral semantic unit.
\textbf{Third}, though the patterns like ``\cspolitician was elected as the president of \cscountry'' are relatively long and rare, they can be grouped with their synonymous patterns so that all the extractions about one entity-attribute type can be aggregated into one set.
That is why \textsf{MetaPAD} successfully discovers who is/was the president of a small country like Burkina Faso or the ceo of a young company like Afghan Citadel.
\textbf{Fourth}, \textsf{MetaPAD} discovered a rich collection of \textsf{person:date\_of\_birth} information from the new corpus that \textit{does not often exist in the knowledge bases}, thanks to our meta patterns use not only entity types but also data types like \csmonth \csday \csyear.

Table~\ref{tab:discoverycvd} shows our \textsf{MetaPAD} can also discover synonymous meta pattern groups and extractions from the biomedical domain. Without heavy annotation of specific domain knowledge, we can find all the patterns about what \cstreatment can treat what \csdisease and what \csbacteria are resistant to what \csantibiotics.

Figure~\ref{fig:patty} shows the SOL pattern synsets that \textsf{PATTY} generates from the four sentences. First, the dependency path loses the rich context around the entities like ``president'' in the first example and ``ceo'' in the last example. Second, the SOL pattern synset cannot group truly synonymous typed textual patterns. We can see the advantages of generating meta patterns and grouping them into synonymous clusters. In the introduction section we also show our \textsf{MetaPAD} can find meta patterns of rich data types for the attribute types like \textsf{person:age} and \textsf{person:date\_of\_birth}.

\subsection{Results on EAV-Tuple Extraction}

Besides directly comparisons on the quality of mining synonymous typed textual patterns, we apply patterns from different systems, \textsf{Ollie} \cite{schmitz2012open}, \textsf{ReNoun} \cite{yahya2014renoun}, and \textsf{PATTY} \cite{nakashole2012patty}, to extract tuple information from the two general corpora \textsf{APR} (news) and \textsf{TWT} (tweets). We attempt to provide quantitative analysis on the use of the typed textual patterns by evaluating how well they can facilitate the tuple extraction which is similar with one of the most challenging NLP tasks called \textit{slot filling} for new attributes \cite{ji2010overview}.

\begin{table}
\vspace{-0.1in}
\caption{Reporting F1, AUC, and number of true positives (TP) on tuple extraction from news and tweets data.}
\vspace{-0.15in}
\Scale[0.88]{\begin{tabular}{l|ccr|ccr}
\toprule
 & \multicolumn{3}{c|}{\textsf{APR} (news, 199MB)} & \multicolumn{3}{|c}{\textsf{TWT} (tweets, 1.05GB)} \\
 & F1 & AUC & TP & F1 & AUC & TP \\ \hline
\textsf{Ollie} \cite{schmitz2012open} & 0.0353 & 0.0133 & 288 & 0.0094 & 0.0012 & 115 \\
\textsf{ReNoun} \cite{yahya2014renoun} & 0.1309 & 0.0900 & 562 & 0.0821 & 0.0347 & 698 \\
\textsf{PATTY} \cite{nakashole2012patty} & 0.3085 & 0.2497 & 860 & 0.2029 & 0.1256 & 860 \\ \hline
\textsf{MetaPAD-T} & 0.3614 & 0.2843 & 799 & 0.3621 & 0.2641 & 880 \\
\textbf{\textsf{MetaPAD-TS}} & 0.4156 & 0.3269 & \textbf{1,355} & \textbf{0.4153} & \textbf{0.3554} & \textbf{1,111} \\
\textsf{MetaPAD-B} & 0.3684 & 0.3186 & 787 & 0.3228 & 0.2704 & 650 \\
\textbf{\textsf{MetaPAD-BS}} & \textbf{0.4236} & \textbf{0.3525} & 1,040 & 0.3827 & 0.3408 & 975 \\
\bottomrule
\end{tabular}}
\label{tab:tupleextraction}
\end{table}

\begin{figure}
\vspace{-0.15in}
\includegraphics[width=3.5in]{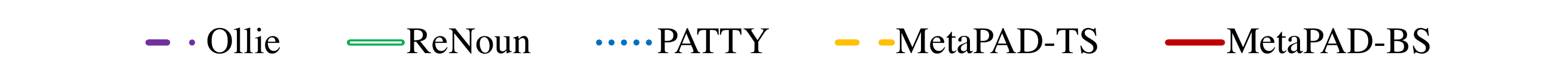} \\
\vspace{-0.1in}
\subfigure[{\textsf{APR} (news, 199MB)}]{\includegraphics[width=1.6in]{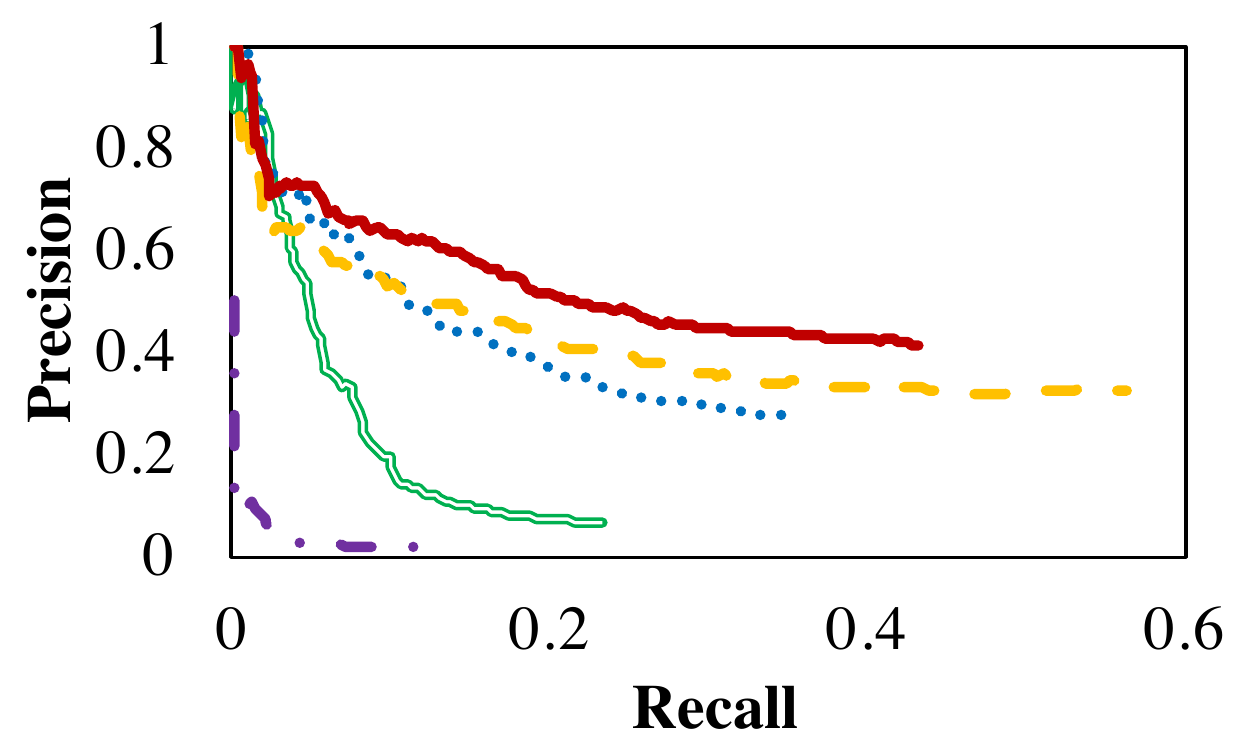}}
\subfigure[{\textsf{TWT} (tweets, 1.05GB)}]{\includegraphics[width=1.6in]{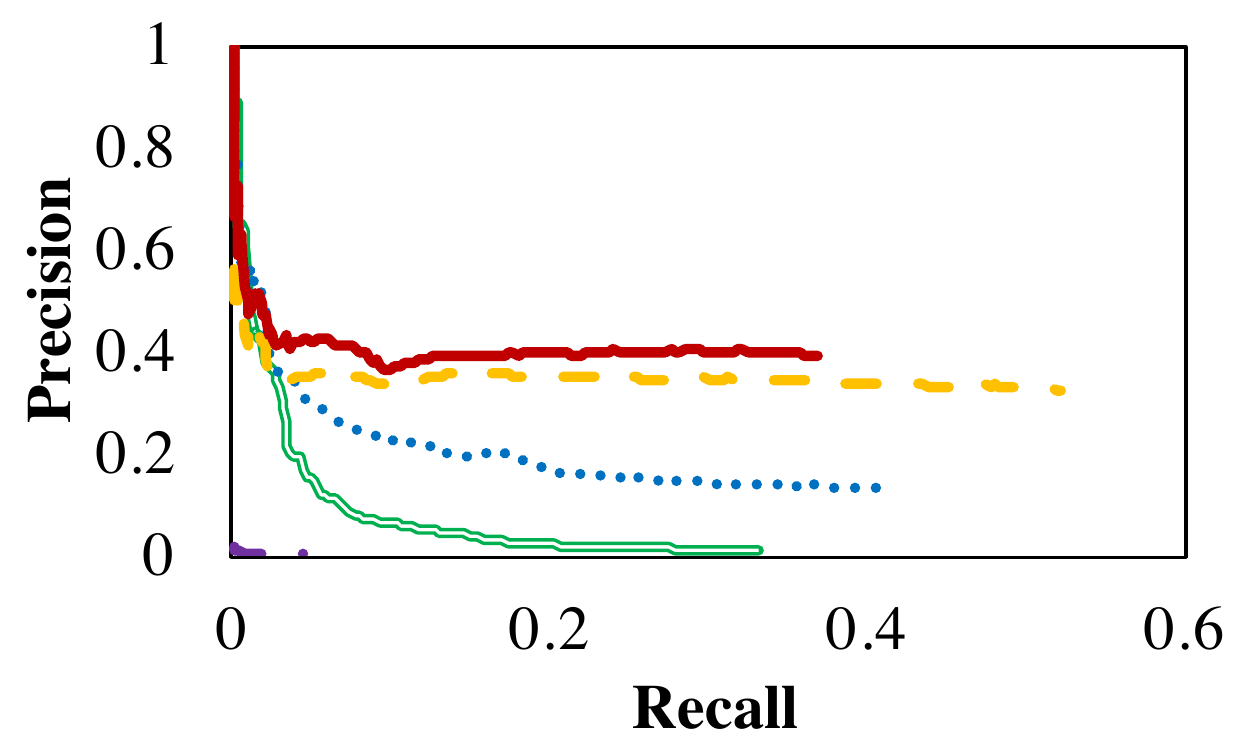}}
\vspace{-0.2in}
\caption{Precision-recall on tuple information extraction.}
\label{fig:valueprcurve}
\vspace{-0.2in}
\end{figure}

\begin{figure*}
\vspace{-0.1in}
\includegraphics[width=7in]{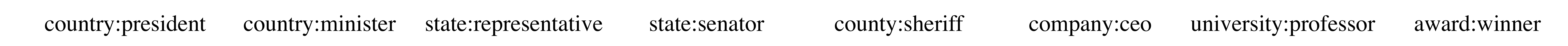} \\
\includegraphics[width=0.14in]{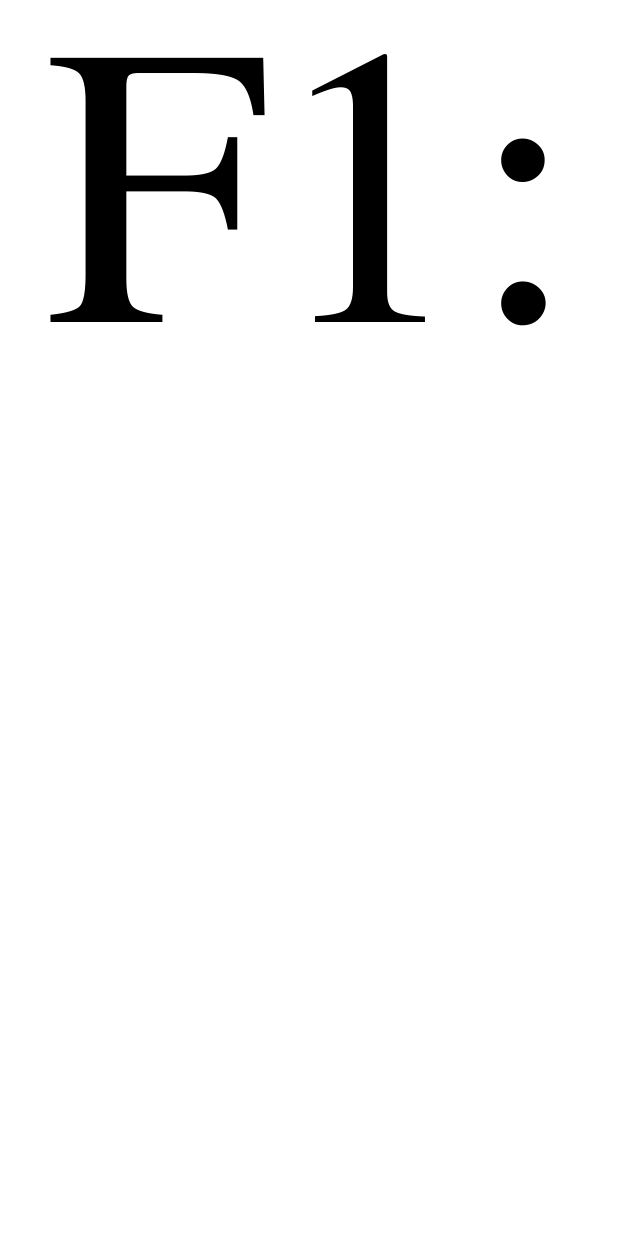}
\includegraphics[width=0.83in]{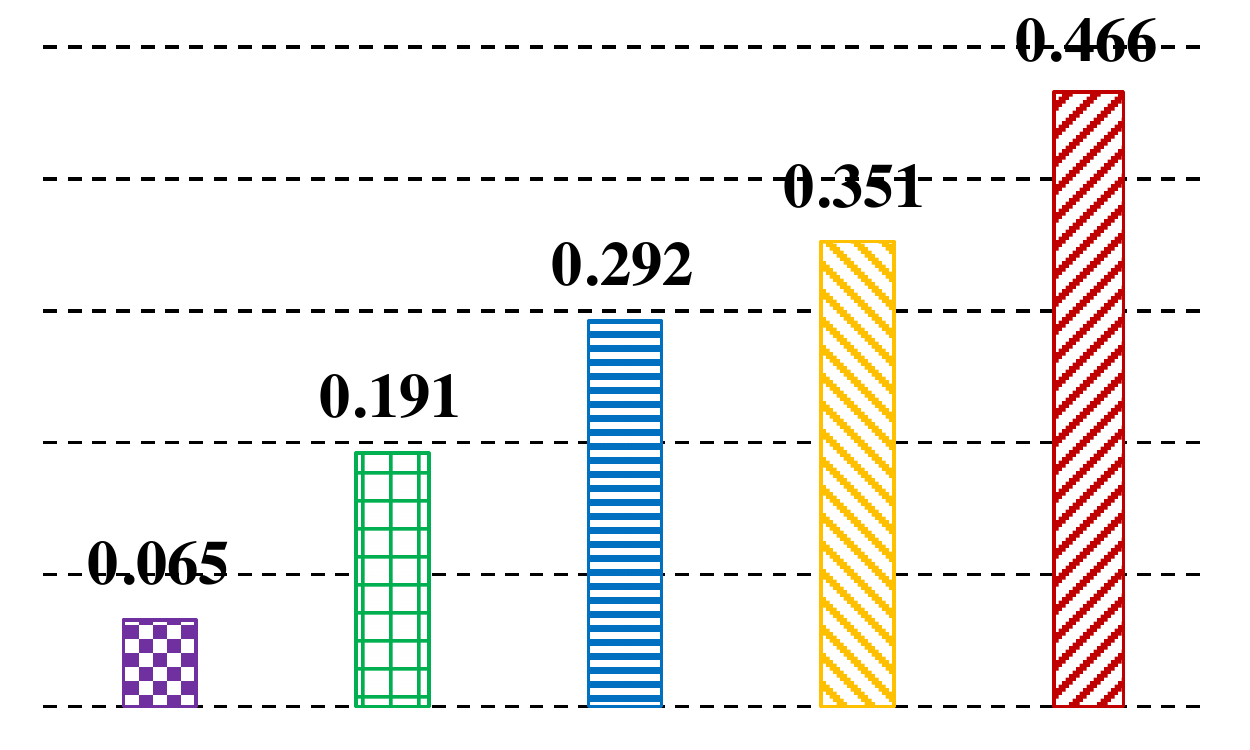}
\includegraphics[width=0.83in]{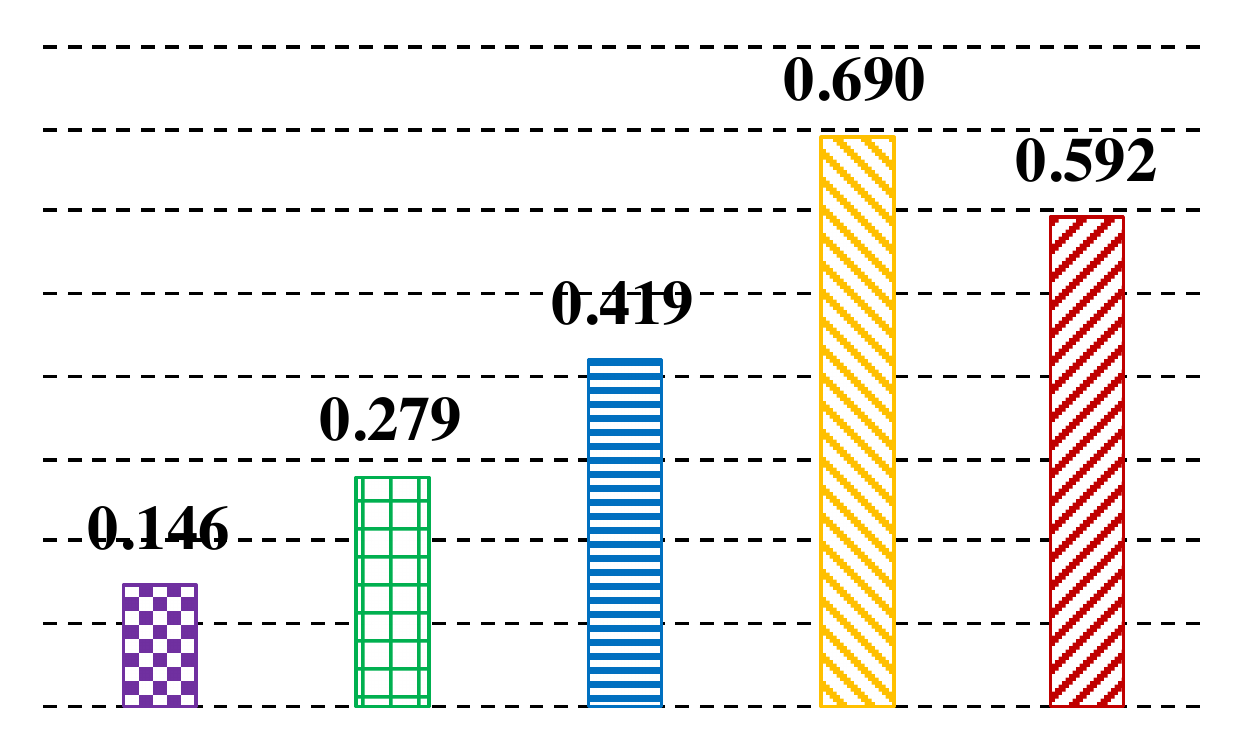}
\includegraphics[width=0.83in]{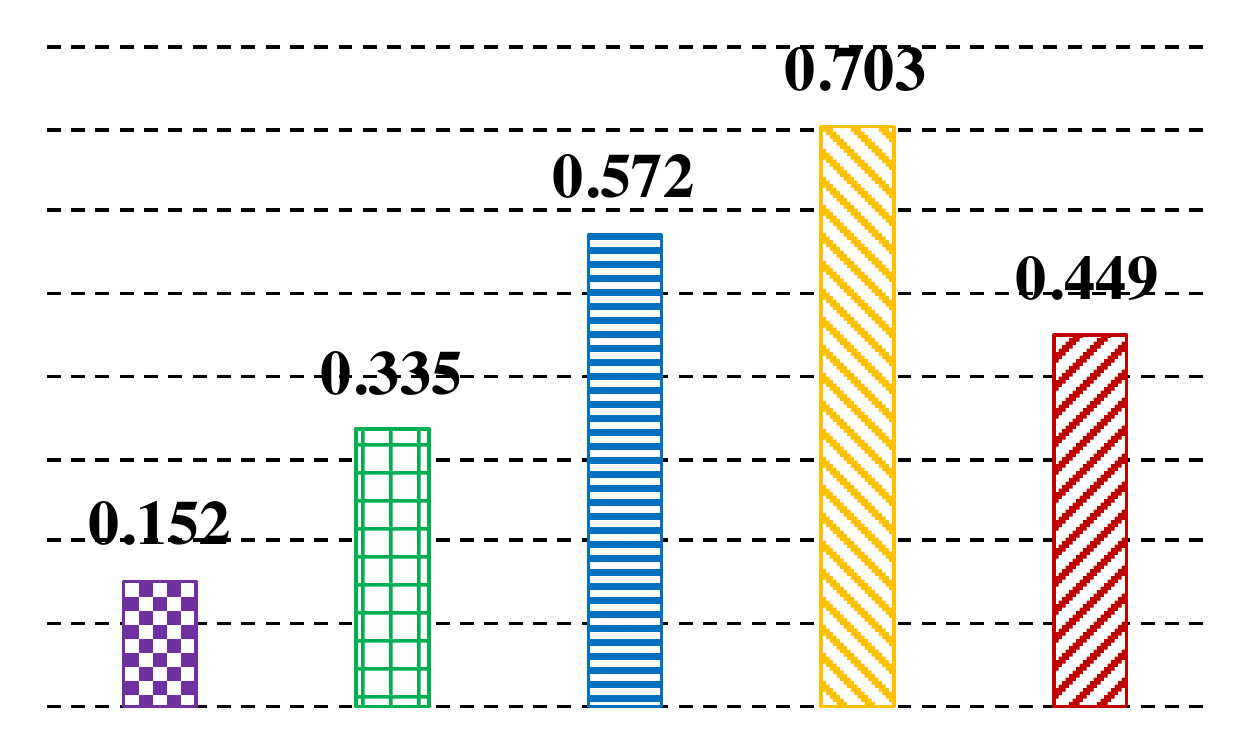}
\includegraphics[width=0.83in]{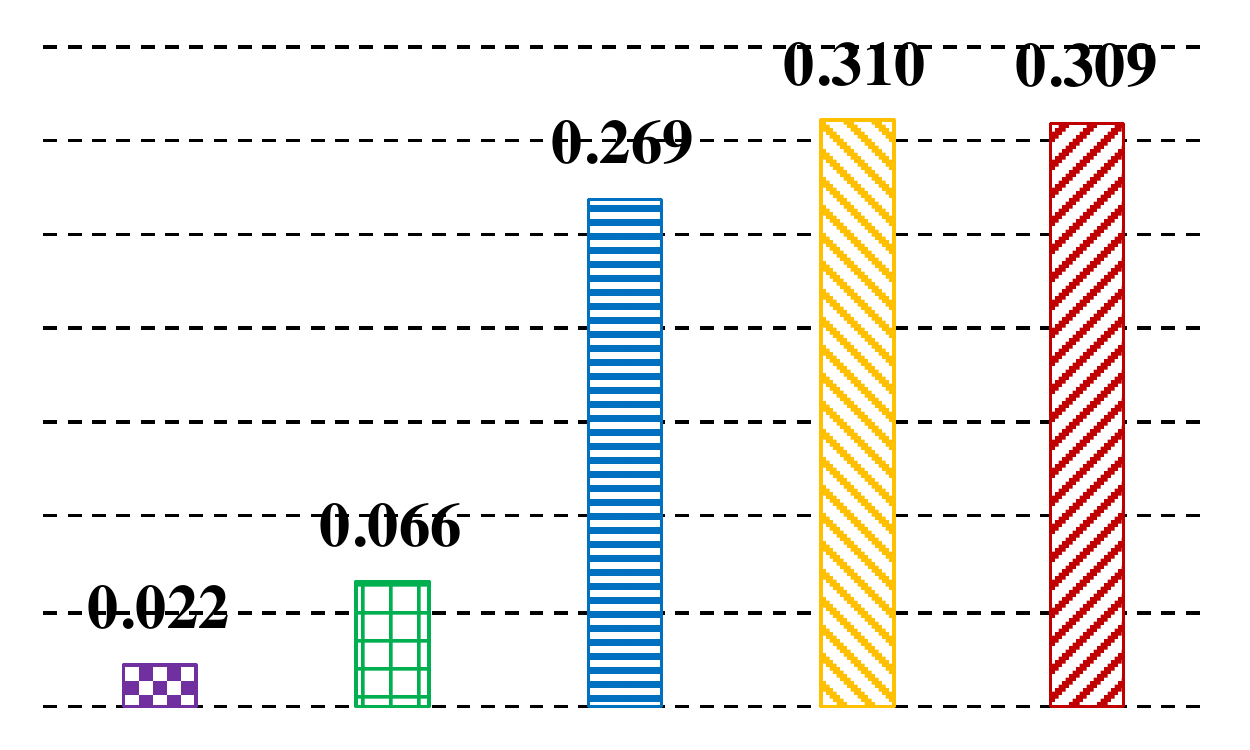}
\includegraphics[width=0.83in]{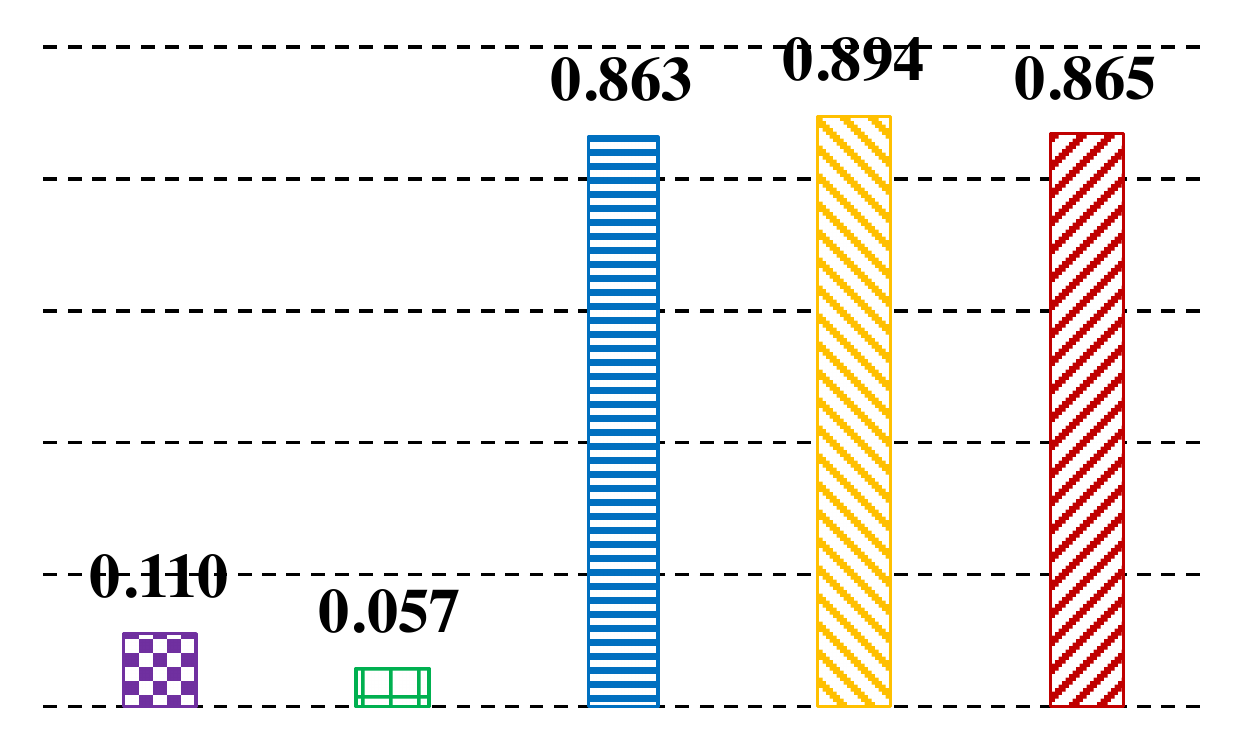}
\includegraphics[width=0.83in]{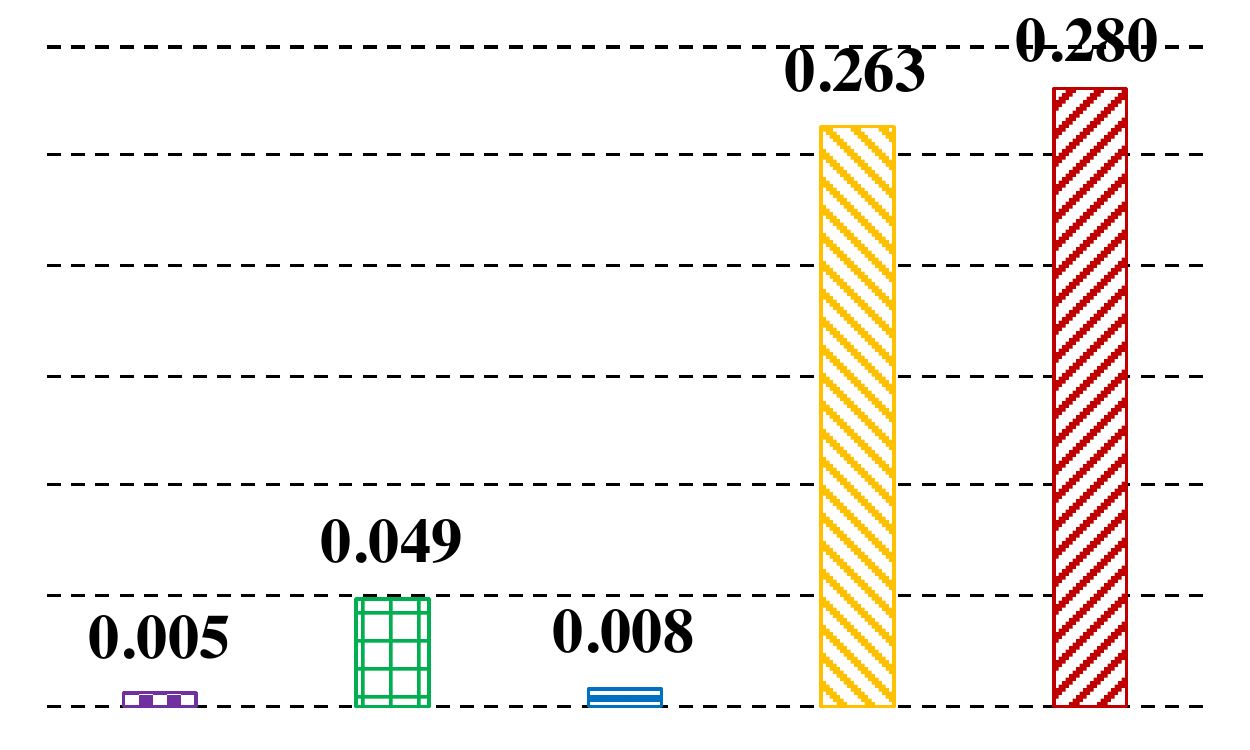}
\includegraphics[width=0.83in]{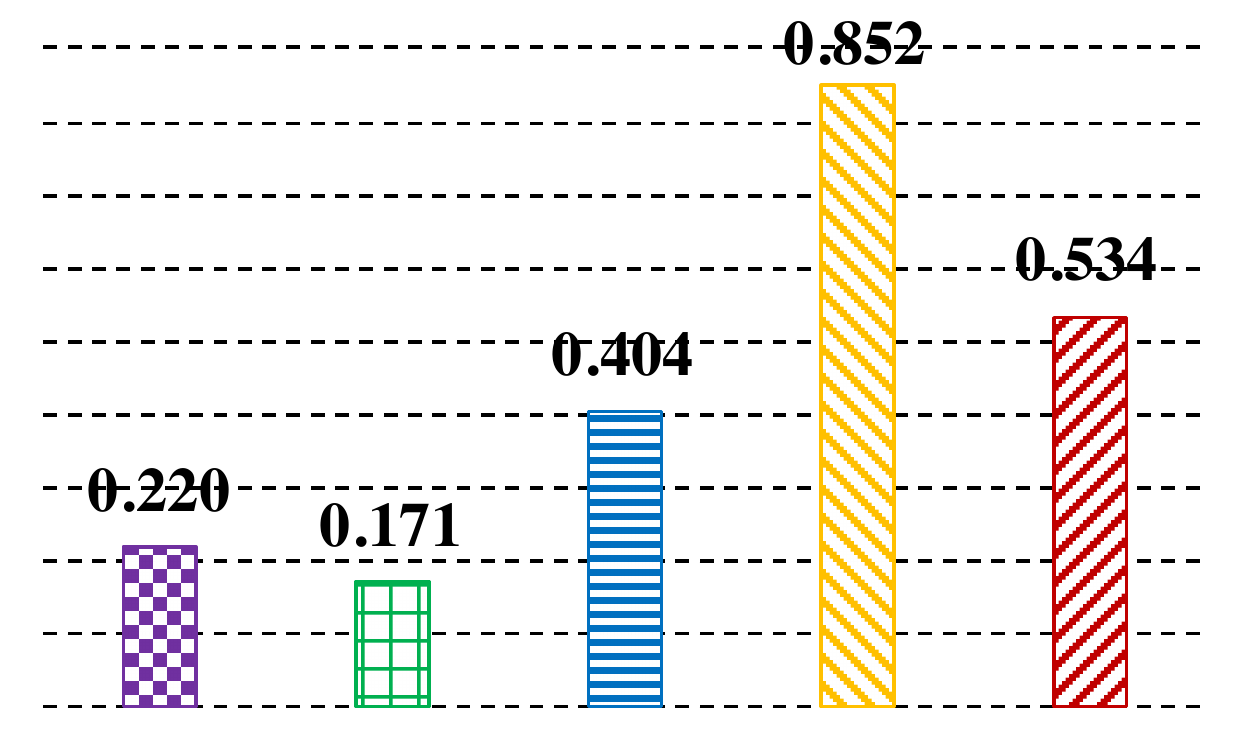}
\includegraphics[width=0.83in]{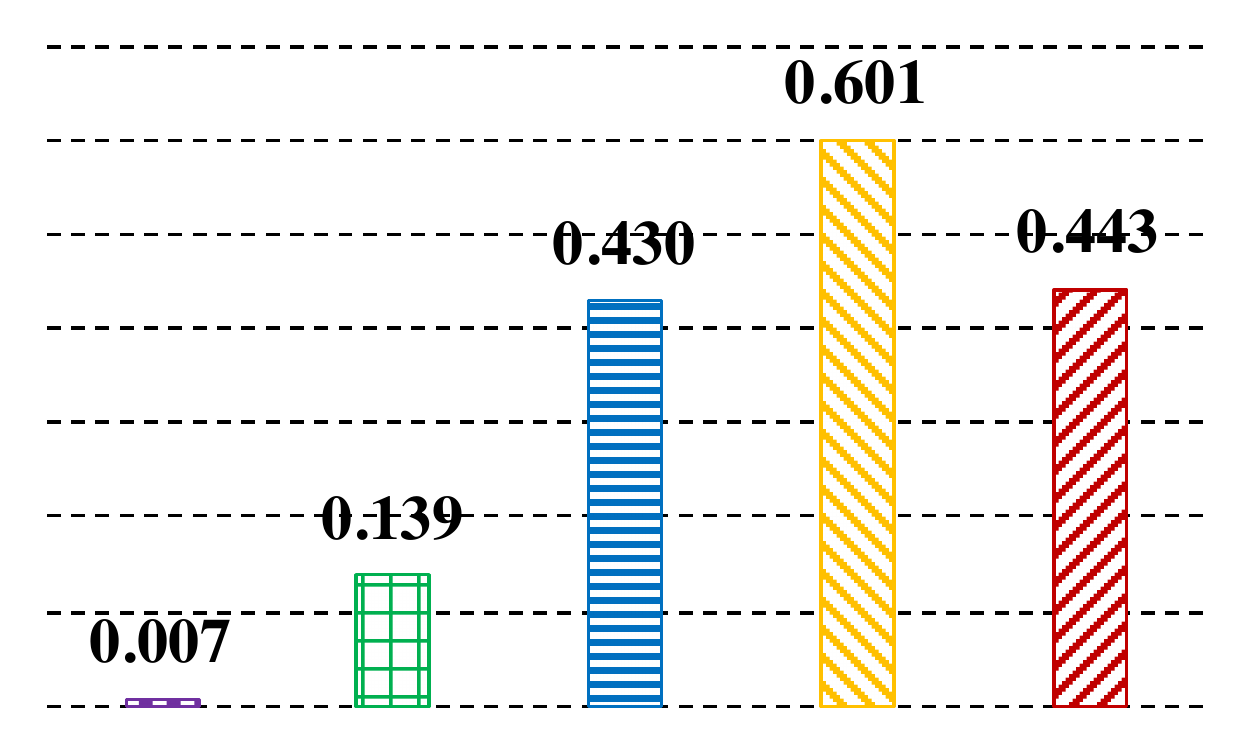} \\
\includegraphics[width=0.14in]{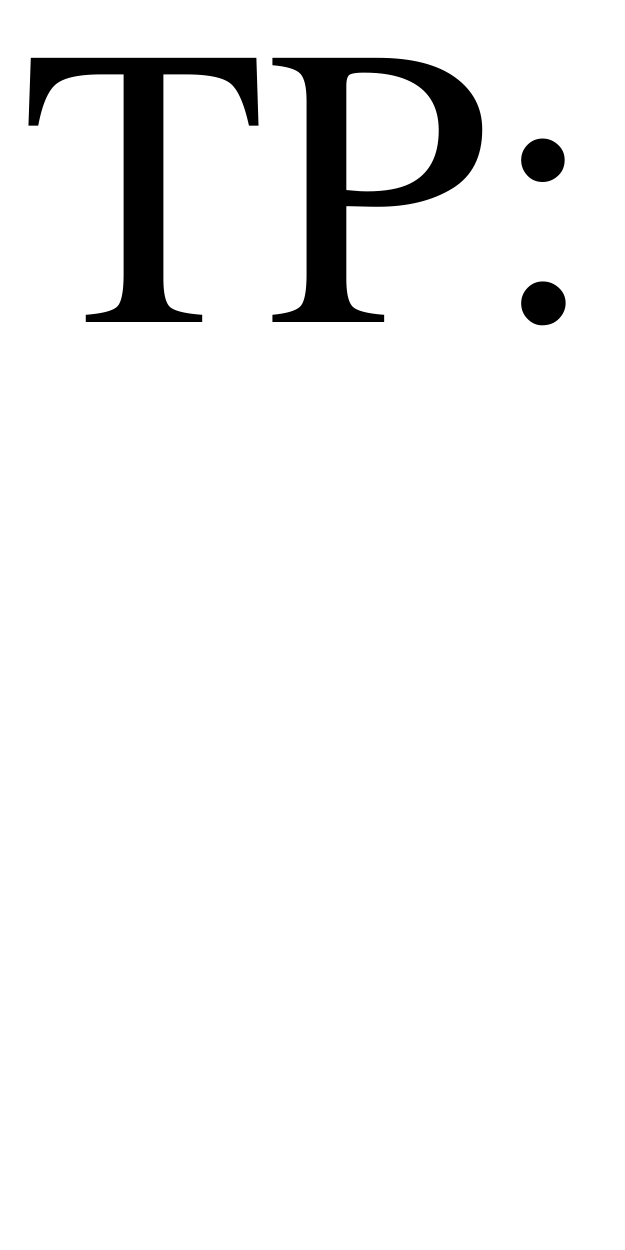}
\includegraphics[width=0.83in]{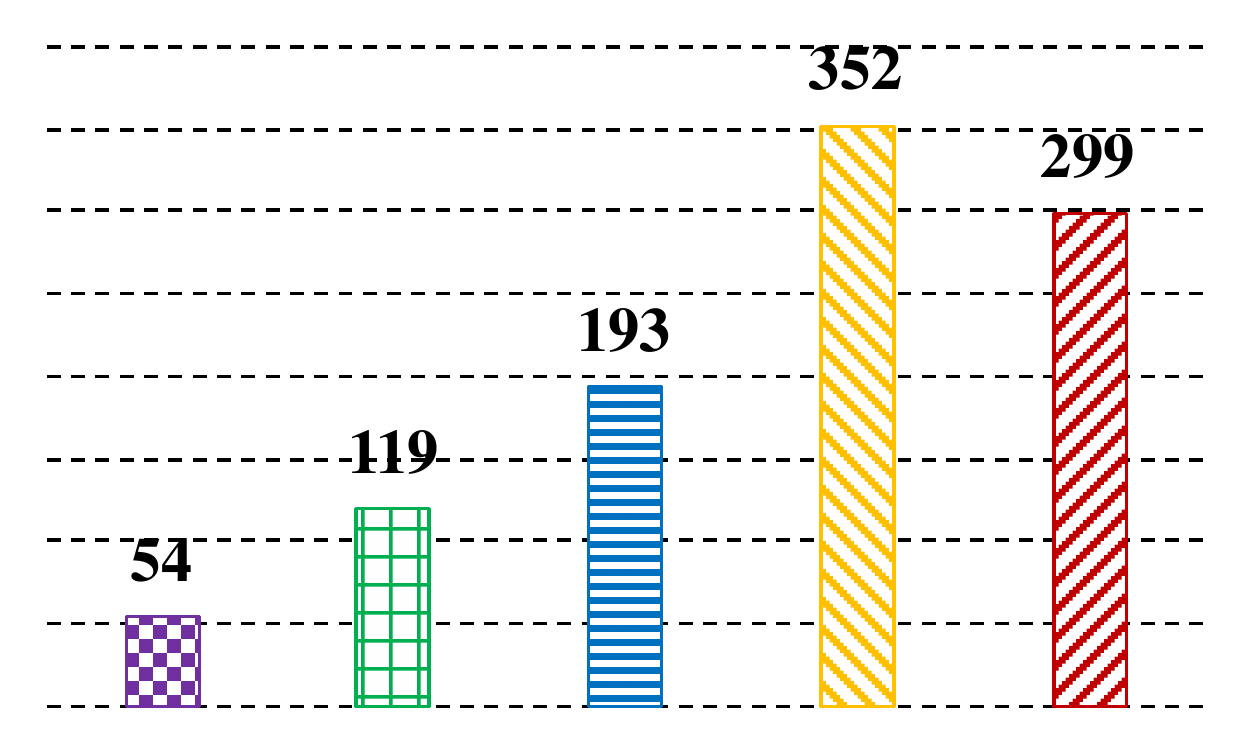}
\includegraphics[width=0.83in]{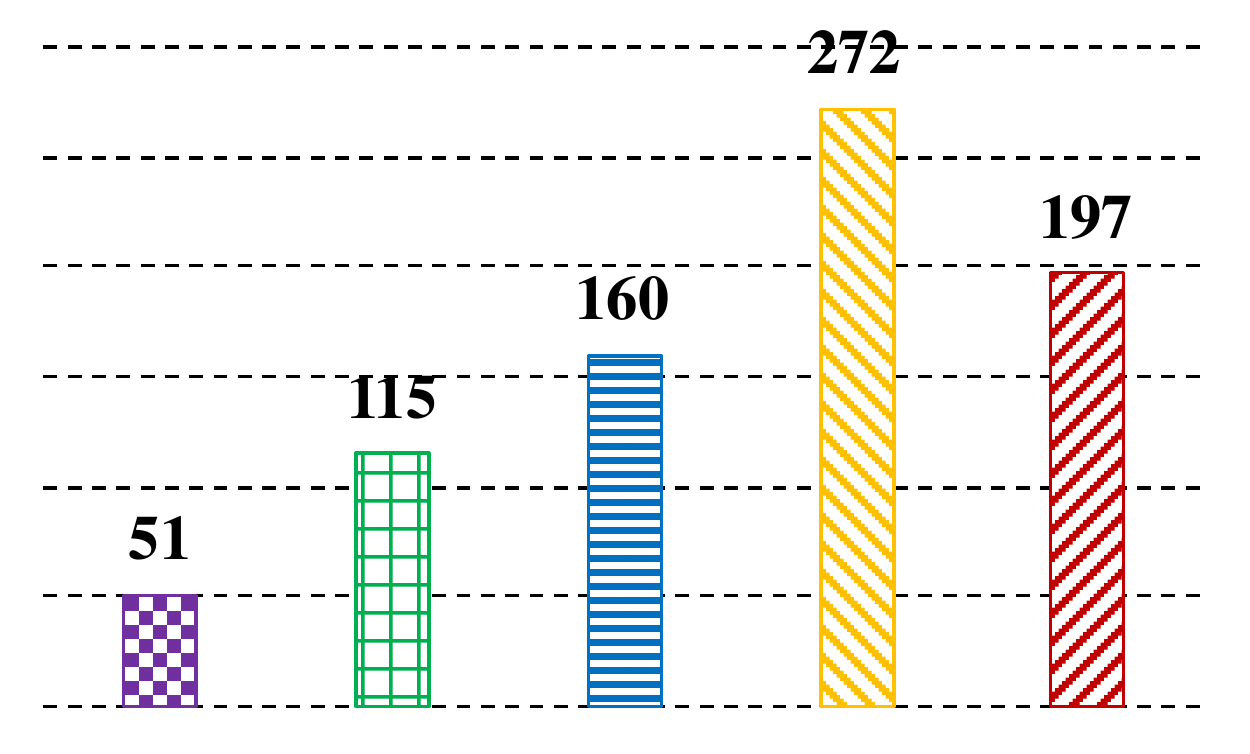}
\includegraphics[width=0.83in]{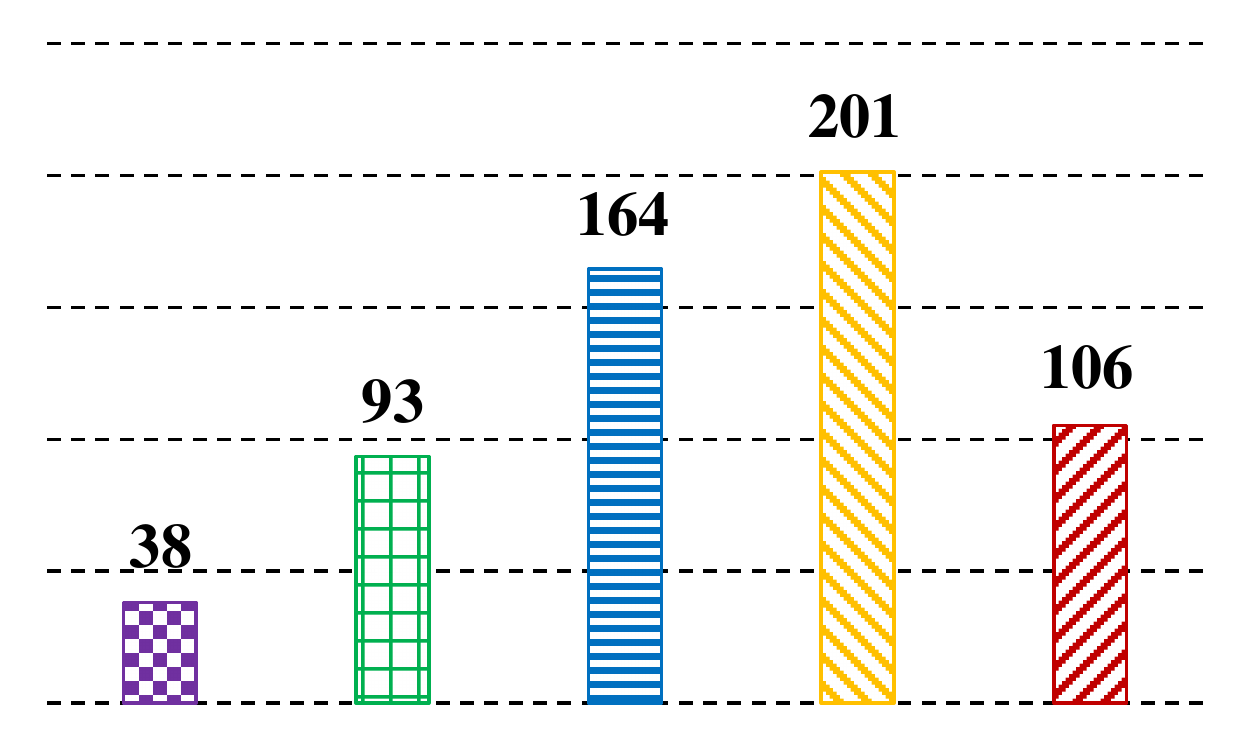}
\includegraphics[width=0.83in]{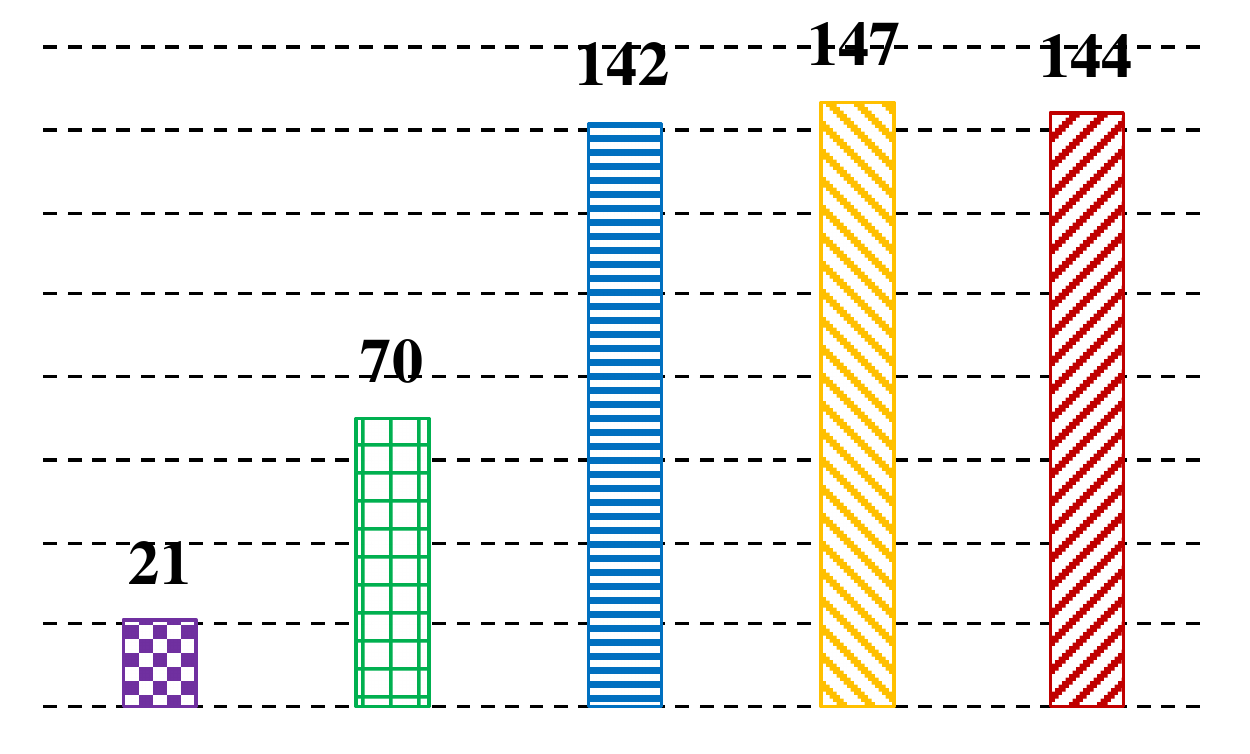}
\includegraphics[width=0.83in]{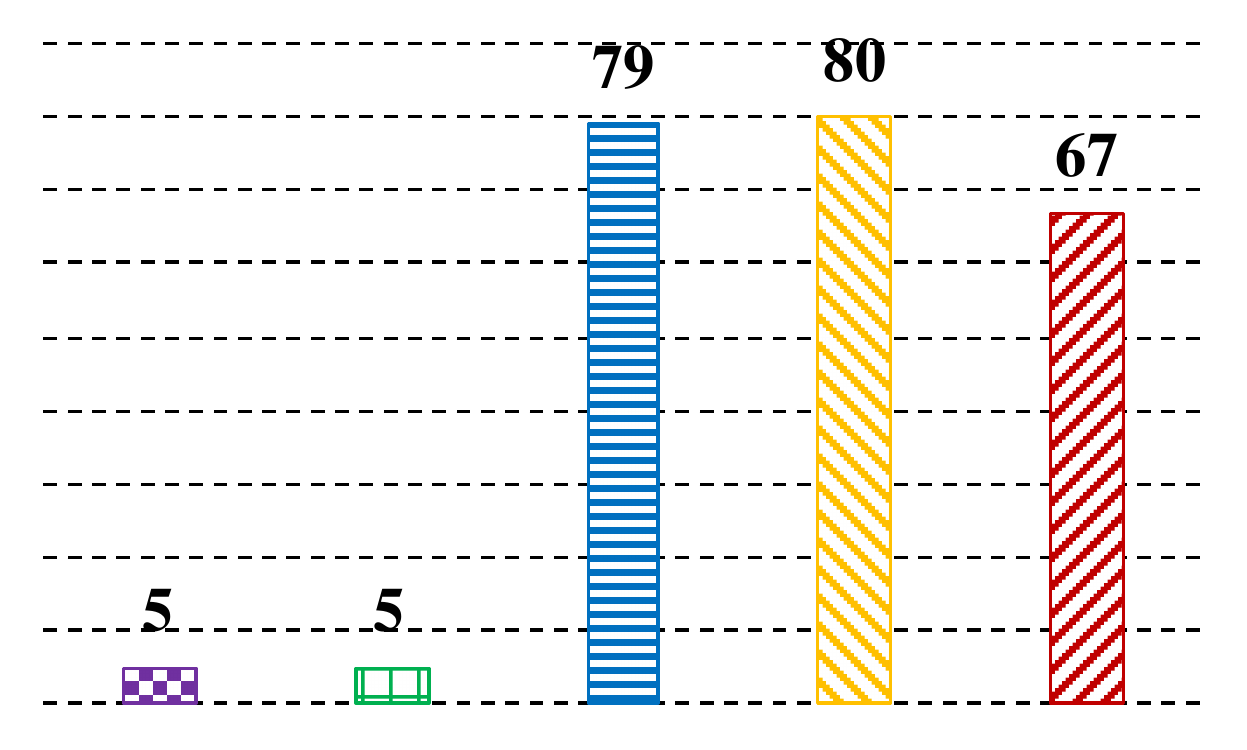}
\includegraphics[width=0.83in]{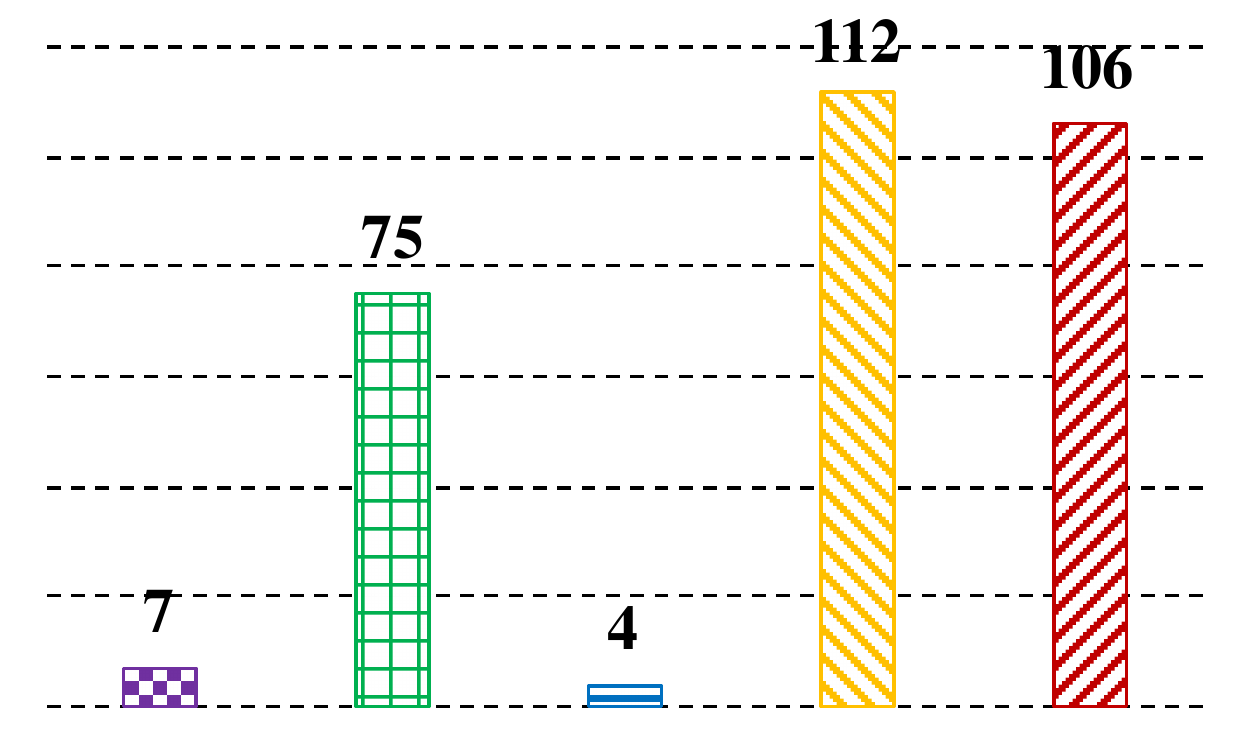}
\includegraphics[width=0.83in]{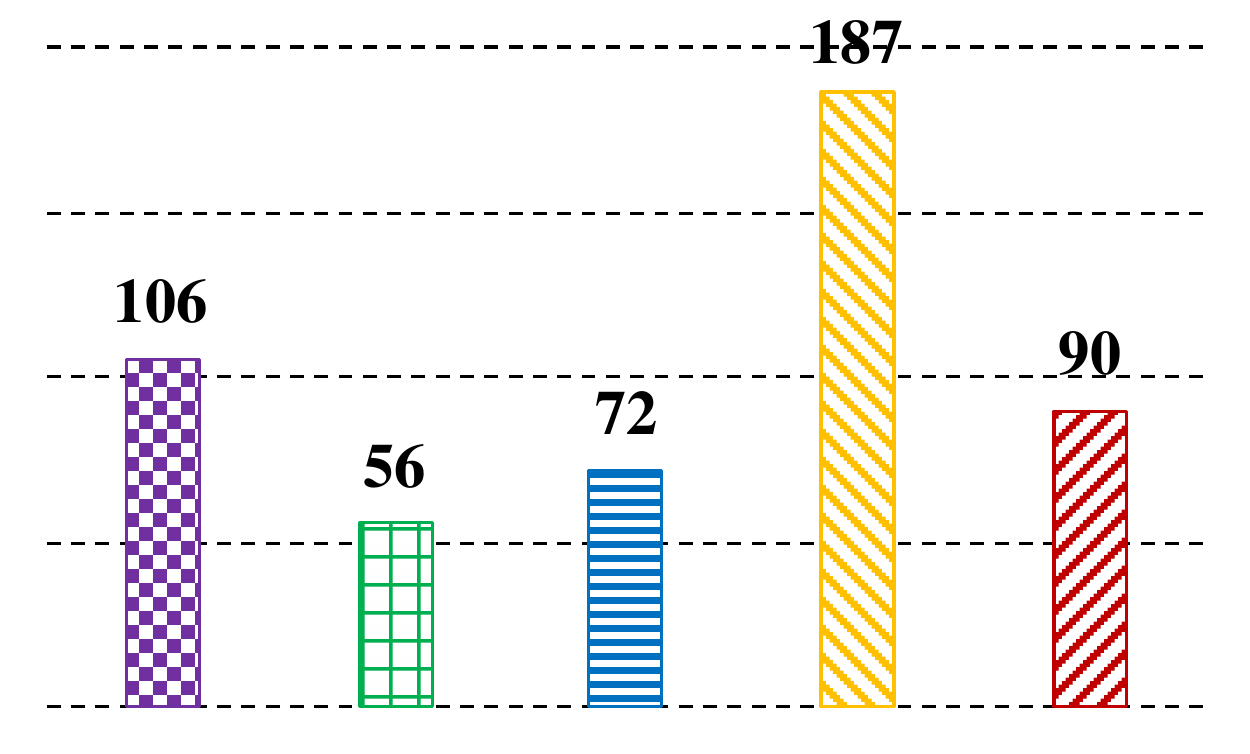}
\includegraphics[width=0.83in]{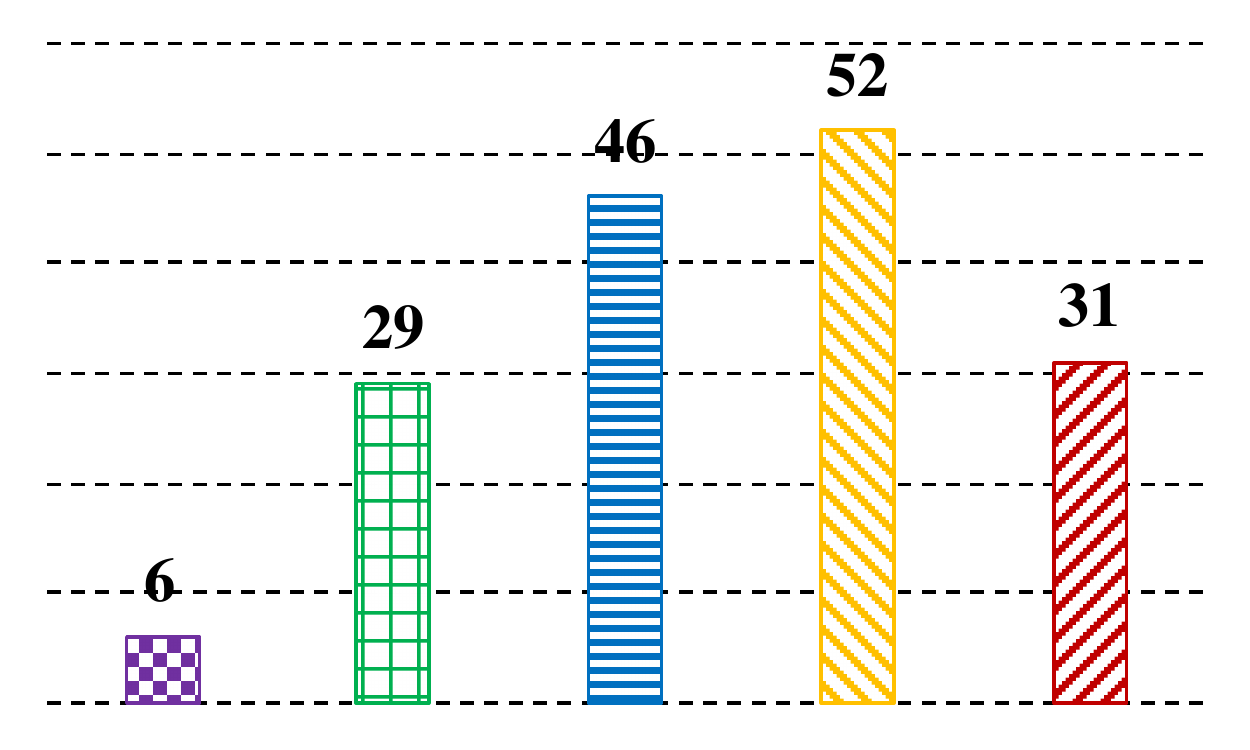} \\
\includegraphics[width=3.8in]{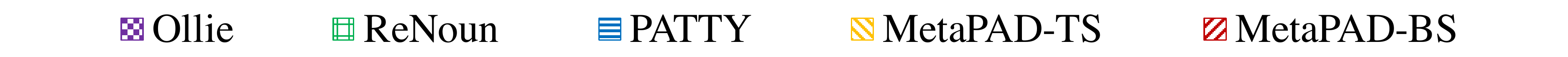}
\vspace{-0.15in}
\caption{Performance comparisons on concrete attribute types in terms of F1 score and number of true positives.}
\label{fig:valuetypebin}
\vspace{-0.15in}
\end{figure*}

Table~\ref{tab:tupleextraction} summarizes comparison results on tuple information that each texutal pattern-driven system extracts from news and tweet datasets. Figure~\ref{fig:valueprcurve} presents precision-recall curves that further demonstrate the effectiveness of our \textsf{MetaPAD} methods. We provide our observation and analysis as follows.

\noindent \textbf{1)} Overall, our \textsf{MetaPAD-TS} and \textsf{MetaPAD-BS} outperform the baseline methods, achieving significant improvement on both datasets (e.g., relatively 37.3\% and 41.2\% on F1 and AUC in the \textsf{APR} data). \textsf{MetaPAD} achieves 0.38--0.42 F1 score on discovering the EAV-tuples of new attributes like \textsf{country:president} and \textsf{company:ceo}. In the TAC KBP competition, the best F1 score of extracting values of traditional attributes like \textsf{person:parent} is only 0.3430 \cite{ji2010overview}. \textsf{MetaPAD} can achieve reasonable performance when working on the \textit{new} attributes. \textsf{MetaPAD} also discovers the largest number of true tuples: on both datasets we discover more than a half of the labelled EAV-tuples (1,355/2,400 from \textsf{APR} and 1,111/2,090 from \textsf{TWT}).

\noindent \textbf{2)} The best of \textsf{MetaPAD-T} and \textsf{MetaPAD-B} that only segment but do not group meta patterns can outperform \textsf{PATTY} relatively by 19.4\% (\textsf{APR}) and 78.5\% (\textsf{TWT}) on F1 and by 27.6\% (\textsf{APR}) and 115.3\% (\textsf{TWT}) on AUC. Ollie parses individual sentences for relational tuples in which the relational phrases are often verbal expressions. So Ollie can hardly find exact attribute names from words or phrases of the relational phrases. \textsf{ReNoun}'s $S$-$A$-$O$ patterns like ``$S$'s $A$ $O$'' require human annotations, use too general symbols, and bring too much noise in the extractions. \textsf{PATTY}'s SOL patterns use entity types but ignore rich context around the entities and only keep the short dependency path. Our meta patten mining has context-aware segmentation with pattern quality assessment, which generates high-quality typed textual patterns from the rich context.

\noindent \textbf{3)} In \textsf{MetaPAD-TS} and \textsf{MetaPAD-BS}, we develop the modules of grouping synonymous patterns and adjusting the entity types for appropriate granularity. They improve the F1 score by 14.8\% and 16.8\% over \textsf{MetaPAD-T} and \textsf{MetaPAD-B}, respectively. We can see the number of true positives is significantly improved by aggregating extractions from different but synonymous meta patterns.

\noindent \textbf{4)} On the tweet data, most of the person, location, and organization entities are NOT able to be typed at a fine-grained level. So \textsf{MetaPAD-T(S)} works better than \textsf{MetaPAD-B(S)}. The news data include a large number of entities of fine-grained types like the presidents and CEOs. So \textsf{MetaPAD-B(S)} works better.

Figure~\ref{fig:valuetypebin} shows the performance on different attribute types on \textsf{APR}. \textsf{MetaPAD} outperforms all the other methods on each type. When there are many ways (patterns) of expressing the attributes, such as \textsf{country:president}, \textsf{company:ceo}, and \textsf{award:winner}, \textsf{MetaPAD} gains more aggregated extractions from grouping the synonymous meta patterns. Our \textsf{MetaPAD} can generate more informative and complete patterns than \textsf{PATTY}'s SOL patterns: for \textsf{state:representative}, \textsf{state:senator}, and \textsf{county:sheriff} that may not have many patterns, \textsf{MetaPAD} does not improve the performance much but it still works better than the baselines.

\subsection{Results on Efficiency}

\begin{table}
\vspace{-0.1in}
\caption{Efficiency: time complexity is linear in corpus size.}
\label{tab:efficiency}
\vspace{-0.15in}
\begin{tabular}{lrrr}
\toprule
& \textsf{APR} & \textsf{CVD} & \textsf{TWT} \\ \hline
File Size & 199MB & 424MB & 1.05GB  \\
\#Meta Pattern & 19,034 & 41,539 & 156,338 \\
Time Cost & 29min & 72min & 117min \\
\bottomrule
\end{tabular}
\vspace{-0.25in}
\end{table}

The execution time experiments were all conducted on a machine with 20 cores of Intel(R) Xeon(R) CPU E5-2680 v2 @ 2.80GHz. Our framework is implemented in \textsf{C++} for \textit{meta-pattern segmentation} and in \textsf{Python} for \textit{grouping synonymous meta patterns} and \textit{adjusting type levels}. We set up 10 threads for \textsf{MetaPAD} as well as all baseline methods. Table~\ref{tab:efficiency} presents the efficiency performance of \textsf{MetaPAD} on three datasets: both the number of meta patterns and time complexity are linear to the corpus size. Specifically, for the 31G tweet data, \textsf{MetaPAD} takes less than 2 hours, while \textsf{PATTY} that requires Stanford parser takes 7.3 hours, and \textsf{Ollie} takes 28.4 hours. Note that for the smaller news data that have many long sentences, \textsf{PATTY} takes even more time, 10.1 hours.

\section{Conclusions}
\label{sec:conclusions}
In this work, we proposed a novel typed textual pattern structure, called \textit{meta pattern}, which is extened to a frequent, complete, informative, and precise subsequence pattern in certain context, compared with the SOL pattern. We developed an efficient framework, \textsf{MetaPAD}, to discover the meta patterns from massive corpora with three techniques, including (1) a context-aware segmentation method to carefully determine the boundaries of the patterns with a learnt pattern quality assessment function, which avoids costly dependency parsing and generates high-quality patterns, (2) a clustering method to group synonymous meta patterns with integrated information of types, context, and instances, and (3) top-down and bottom-up schemes to adjust the levels of entity types in the meta patterns by examining the type distributions of entities in the instances. Experiments demonstrated that \textsf{MetaPAD} efficiently discovered a large collection of high-quality typed textual patterns to facilitate challenging NLP tasks like tuple information extraction.

\bibliographystyle{ACM-Reference-Format}
\bibliography{paper}

\end{document}